\documentclass[12pt]{article}

\usepackage{cite}
\usepackage{amsmath,amssymb,amsfonts}
\usepackage{algorithmic}
\usepackage{textcomp}
\usepackage{rotating} 
\usepackage[utf8]{inputenc} 
\usepackage[T1]{fontenc}    
\usepackage{hyperref}       
\usepackage{url}            
\usepackage{booktabs}       
\usepackage{amsfonts}       
\usepackage{nicefrac}       
\usepackage{microtype}      
\usepackage{lipsum}		
\usepackage{graphicx}
\usepackage{natbib}
\usepackage{multirow}
\usepackage[acronym]{glossaries}
\usepackage{rotating}
\usepackage{subcaption}
\usepackage{float}
\usepackage{xcolor}
\usepackage{lineno}

\newacronym{semg}{sEMG}{surface electromyography}
\newacronym{emg}{EMG}{electromyography}
\newacronym{mnf}{MNF}{mean frequency}
\newacronym{mav}{MAV}{mean absolute value}
\newacronym{iemg}{iEMG}{integrated EMG}
\newacronym{ecg}{ECG}{Electrocardiography}
\newacronym{eda}{EDA}{electrodermal activity}
\newacronym{ppg}{PPG}{photoplethysmography}
\newacronym{imu}{IMU}{Inertial Measurement Unit}
\newacronym{hrv}{HRV}{hear rate variability}
\newacronym{fds}{FDS}{flexor digitorum superficialis}
\newacronym{stai}{STAI}{State-Trait Anxiety Inventory}
\newacronym{rnn}{RNN}{Recurrent Neural Network}
\newacronym{hr}{HR}{heart rate}
\newacronym{gru}{GRU}{Gated Recurrent Unit}
\newacronym{mlp}{MLP}{Multilayer Perceptron}
\newacronym{cnn}{CNN}{Convolutional Neural Network}
\newacronym{mse}{MSE}{mean squared error}
\newacronym{rmse}{RMSE}{root mean squared error}
\newacronym{mae}{MAE}{Mean absolute error}
\newacronym{iid}{i.i.d.}{independent and identical distributed}
\newacronym{muap}{MUAP}{motor unit action potential}
\newacronym{muapt}{MUAPT}{motor unit action potential train}
\newacronym{cns}{CNS}{central nervous system}
\newacronym{ans}{ANS}{autonomic nervous system}
\newacronym{eeg}{EEG}{electroencephalogram}
\newacronym{rmssd}{RMSSD}{root mean square of successive differences}
\newacronym{lstm}{LSTM}{Long short-term memory}

\begin{document}

\title{The Climber's Grip - Personalized Deep Learning Models for Fear and Muscle Activity in Climbing}
\author{Matthias Boeker$^{1}$, Dana Swarbrick$^{2,3}$, Ulysse T.A. Côté-Allard$^{4}$,\\Marc T.P. Adam$^{5}$, Hugo L. Hammer$^{6}$, and Pål Halvorsen$^{1}$}
 \date{}

\maketitle

\paragraph*{Keywords:} Fear, Climbing, Deep learning, Random Effects, Time series analysis

\begin{abstract}
Climbing is a multifaceted sport that combines physical demands and emotional and cognitive challenges. Ascent styles differ in fall distance with \textit{lead climbing} involving larger falls than \textit{top rope climbing}, which may result in different perceived risk and fear. In this study, we investigated the psychophysiological relationship between perceived fear and muscle activity in climbers using a combination of statistical modeling and deep learning techniques. We conducted an experiment with 19 climbers, collecting electromyography (EMG), electrocardiography (ECG) and arm motion data during \textit{lead} and \textit{top rope climbing}. Perceived fear ratings were collected for the different phases of the climb. Using a linear mixed-effects model, we analyzed the relationships between perceived fear and physiological measures. To capture the non-linear dynamics of this relationship, we extended our analysis to deep learning models and integrated random effects for a personalized modeling approach. Our results showed that random effects improved model performance of the mean squared error (MSE), mean absolute error (MAE) and root mean squared error (RMSE). The results showed that muscle fatigue correlates significantly with increased fear during \textit{lead climbing}. This study highlights the potential of combining statistical and deep learning approaches for modeling the interplay between psychological and physiological states during climbing.
\end{abstract}


\maketitle

\section{Introduction}\label{sec:introduction}
Climbing is a physically demanding sport in which climbers must carefully manage their physical exertion throughout an ascent. In addition to physical demands, climbing requires cognitive and emotional control \cite{hodgson2009perceived}. There are different types of climbing ascent styles related to different safety rope protocols that differ in their psychophysiological demands, at least for intermediate climbers \cite{hodgson2009perceived,draper2010physiological, mangan2024psychology}. In this study, we investigated the ascent styles of \textit{top rope} and \textit{lead climbing}. In \textit{top rope} climbing, the rope is anchored above the climber, which minimizes the fall distances and perceived risk. Conversely, in \textit{lead climbing}, the climber must manage and secure the rope to points of protection along the route. Therefore, they are confronted with variable and often longer fall distances, which may elicit a higher level of fear \cite{giles2014current, pijpers2003anxiety, pijpers2005anxiety}. While climbers widely acknowledge the interplay between the experience of fear and physical exertion, the relationship between fear and physiological responses, such as muscle activity and muscle fatigue, has not yet been rigorously investigated. 
To address this gap, we conducted an experiment to explore the relationship between self-reported fear and muscle activity among climbers.
In this experiment, we tracked the physiological signals of 20 recreational climbers on a 19.3 m high indoor climbing wall. We measured the subjective experience of fear using self-reports and muscle activity through \gls{emg} of the forearm. The climbers engaged in both \textit{top rope} and \textit{lead climbing} in a within-subjects design. We investigated the impact of height and ascent styles on fear. Furthermore, we examined the relation between self-reported fear, muscle activity, and fatigue.
A common statistical approach to test linear relationships in data collected from different participants is linear mixed effects modeling \cite{laird1982random}. To account for individual differences in climbing experience, mental resilience, and physical capacity, we included participants as random effects to capture variability across individuals. This approach enabled us to examine the psychophysiological influences of different climbing styles and physiological responses on subjective fear.
In contrast to the traditional approach of linear mixed effects modeling, we also explored the application of more complex deep learning methods. These algorithms have the potential to recognize complex patterns within signals and to model non-linear relations. We model the physiological signals with a \gls{gru} deep learning algorithm. The \gls{gru} architecture is able to capture temporal dependencies in sequential data. Unlike simpler models, \gls{gru} maintain an internal memory state that allows them to effectively learn patterns across time steps without suffering from vanishing gradient problems common in traditional recurrent networks \cite{chung2014empirical}. We compare its performance with \gls{mlp} and a linear model as benchmark.
To account for individual differences between the participants, similar to the linear mixed effects model, we integrated random effects into the deep learning architectures, following Simchoni et al. \cite{simchoni2023integrating}. In general, deep learning models are population-based, meaning that one model is trained on all individuals. 
Mixed effects deep learning architectures allow for a more personalized modeling approach \cite{chung2020deep}. The inclusion of random effects in deep learning has only recently been explored in the literature by  Simchoni and Rosset and Kilian et al.\cite{simchoni2021using, simchoni2023integrating, kilian2023mixed}. Our models showed that the inclusion of random effects to account for individual differences between climbers often improved model performance as measured with \gls{mse}, \gls{mae} and \gls{rmse}. 

In summary, the main contributions are as follows.
\begin{enumerate}
    \item In Sport Psychology.
    \begin{enumerate}
        \item We conducted a within-subjects experiment and collected multimodal data, including physiology with sensors, behavior with video recording, and subjective experience with questionnaires.
        \item We demonstrated a significant relationship between the subjective experience of fear and muscle activity, highlighting the interplay between psychological and physiological responses.
        \item Specifically, muscle fatigue was more strongly associated with perceived
fear during lead climbing than top rope climbing.
    \end{enumerate}
    \item Field of statistical machine learning.
    \begin{enumerate}
        \item We implemented and evaluated the integration of random effects in deep learning, enabling personalized modeling of our dataset.
        \item We showed that by accounting for random effects between the climbers, model performances regarding \gls{mse}, \gls{mae} and \gls{rmse} generally improved. 
        \item We provide a PyTorch layer and a loss function for straightforward integration of random effects. 
    \end{enumerate}
\end{enumerate}

\section{Background}
\subsection{Muscle activity and fatigue}\label{sec:muscle_fatigue}
Muscle activity refers to the physiological processes by which muscle fibers are stimulated to contract by motor neurons. A group of muscle fibers is stimulated by the nervous system through motor neurons that sit in the spinal cord \cite{basmajian1962muscles}. This compound is referred to as a motor unit. The electrical signal sent from the terminal branch of the motor neuron axon propagates through the muscle fiber, causing its contraction. The summed electrical signal detected in the group of muscle fibers in the motor unit is referred to as \gls{muap}. A repetitive sequence of \gls{muap} is called a \gls{muapt}. The \gls{emg} signal measures the summation of the \gls{muapt} from all the active motor units within the area covered by the electrodes \cite{basmajian1962muscles}. 
The raw \gls{emg} signal is a complex electrical recording that represents the electrical activity generated by the muscle fibers during contraction. Due to its complexity and the presence of noise, the signal is pre-processed to better interpret and quantify muscle activity. The amplitude of the rectified \gls{emg} signal is related to the level of muscle activity, which can be measured using \gls{iemg} \cite{cifrek2009surface, clancy2002sampling}.

Prolonged or intense muscle activity can lead to muscle fatigue. Muscle fatigue is a localized and complex physiological state characterized by a temporary decrease in the ability to perform physical tasks~\cite{enoka2008muscle, cifrek2009surface}. This condition can reduce the rate at which the motor neurons fire action potentials. This reduced firing rate indicates that \gls{muap} within a \gls{muapt} occurs less frequently. The reduced frequency leads to reduced stimulation of the muscle fibres, which reduces the muscle's ability to generate and maintain a level of activity \cite{constantin2021molecular}. This condition is due to various changes within the muscle that affect metabolism, structure and energy supply and is primarily due to insufficient oxygen and nutrient supply~\cite{cifrek2009surface}. Numerous physiological processes contribute to muscle fatigue, making it difficult to identify the exact causes~\cite{enoka2008muscle}. Muscle contractions increase arterial blood pressure, which in turn reduces blood flow to active muscles. This reduction restricts oxygen supply and contributes to muscle fatigue. Reduced blood flow also hinders the removal of metabolic waste, which leads to the accumulation of lactic acid in the muscle tissue \cite{wan2017muscle}. This accumulation often leads to a ‘burning’ sensation or feeling ‘pumped’ and the muscle appears swollen. In addition, increased acidity reduces the speed at which electrical signals travel through the muscle \cite{cifrek2009surface}.
The most common measure of muscle fatigue is the decrease of \gls{mnf} of the \gls{emg} signal \cite{cifrek2009surface, de1984myoelectrical}. 
An increase in amplitude is also correlated with increased fatigue \cite{maton1981human}. While individual motor units reduce their firing rate due to fatigue, increasing amplitude indicates that additional motor units are recruited to compensate for the loss of contractility \cite{eason1960electromyographic, maton1981human}.

\subsection{Fear during climbing} \label{sec:fear_background}
Climbing is considered an adventure sport because of the exposure to height and risk of falling, both of which contribute to the sense of thrill and excitement. In climbing, the risk of injury from falling can be real or perceived. Generally, indoor climbing offers a safe, controlled environment free from risk as long as proper safety procedures are followed. Nonetheless, climbing can elicit fear of falling, fear of heights, and anxiety in climbers. 

Several previous studies have compared psychophysiological measures between \textit{lead} and \textit{top rope} climbing. Hodgson et al. found that \textit{lead climbing} elicited higher cognitive and somatic stress compared to \textit{top rope climbing} in intermediate climbers \cite{hodgson2009perceived}.  Draper, Fryer, and colleagues found that intermediate climbers reported that leading was more physically and mentally demanding than top roping \cite{draper2010physiological}. However, when examining advanced and elite climbers, while they spent longer on a lead climb than a top-rope climb, there were no statistically significant differences on psychophysiological outcome measures \cite{fryer2012effect, dickson2012effect}. Therefore, the experience level of the climbers is an important factor when considering psychophysiological responses during climbing. These expertise differences are likely related to how more experienced climbers will have been exposed to fewer unsafe lead falls and more safe experiences at heights. A recent review highlights the importance of not conflating between experience level and climbing ability when examining differences in climbing ascent style \cite{mangan2024psychology}. 

Height is another climbing variable that may impact fear. In a series of experiments, Pijpers et al. manipulated the height of identical traverses and measured inexperienced and beginner climbers' psychophysiological responses \cite{pijpers2003anxiety, pijpers2005anxiety, pijpers2006role}. Participants' physiological responses exhibited significantly more muscle fatigue, higher blood lactate concentrations, and higher heart rates on the high route \cite{pijpers2003anxiety}. They also found behavioral differences, including that the higher traverse elicited longer climbing times, less smooth motion, more explosive movements, and slower, prolonged grasps \cite{pijpers2003anxiety, pijpers2005anxiety}. They also found that participants perceived and actual maximal reaching height was reduced in the higher traverse \cite{pijpers2006role}. Therefore, another important variable to consider when examining fear and anxiety in climbing is height. 

There are many theories of emotion and affect that are useful when conceptualizing emotions like fear and anxiety. According to Ekman, fear is a basic emotion \cite{ekman1992there}. However, according to Russell et al.'s circumplex model of affect, fear is not an independent, atomic state but can be conceptualized on a scale of arousal and valence \cite{russell2003core}. In Russell's model, fear is characterized by negative valence and increased arousal \cite{posner2005circumplex}. While valence is a scale of pleasure–displeasure, arousal is the activation of the sympathetic \gls{ans} \cite{russell2003core}. High arousal manifests itself in physiological reactions such as increased heart rate, sweat gland activation, blood flow to muscles, release of adrenaline into the bloodstream, and muscle tension \cite{lebouef2023physiology}. The subjective experience of fear (high arousal, low valence) or excitement (high arousal, high valence) may be the result of the interpretation of these physiological responses, in combination with a state of valence \cite{posner2005circumplex}.
In the literature, the term anxiety is often used interchangeably with the term fear. We have defined both terms to avoid confusion. Parts of the literature argue that anxiety and fear are the same phenomenon \cite{steimer2002biology}. However, a difference can be defined in the objectives of the two terms. While fear is focused on a real, external, and known danger, anxiety is directed at an unknown threat. Both fear and anxiety cause sympathetic arousal that triggers physiological responses \cite{steimer2002biology, craig1995environmental}. In climbing, risk and threats can be real or imagined, so it is challenging to know which term fits best in the climbing literature. We have chosen to use the term "fear" instead of "anxiety" because two of the subjective self-report outcome measures included this term, and only one included anxiety. However, we acknowledge that the field would benefit from a critical review of the usage of affective terminology.

\section{Methods}
First, we describe the climbing experiment and the collection and processing of the psychological and physiological measures. Then, we present methods for modeling random effects in deep learning and model evaluation.

\subsection{Climbing Experiment}
This section describes the methodology of the data collection and processing for the climbing experiment. It is divided into subsections describing the experimental protocol, sensor setup, synchronization process between video and sensor data, and steps for pre-processing and extracting meaningful features from physiological signals.

\subsubsection{Participants}
We conducted a study with 20 recreational climbers, comprising eight women and twelve men, with an average age of 31 years ($\bar{x} = 7.8$, range: $23-50$ years). Participants were required to have a certification from the Norwegian Climbing Association indicating that they could lead belay safely. The participants had varying levels of climbing experience, with climbing experience ranging from 1.5 years to 24 years ($\bar x = 6.6$, $s = 6.84$ years). The bouldering experience ranged from 0 to 20 years ($\bar x = 4.3$, $s = 4.62$ years). Participants dedicated an average of 1.5 hours per week to \textit{lead climbing}, one hour per week to \textit{top rope climbing}, and one hour per week to bouldering. Notably, there was considerable variation in the time spent on these activities, with some participants spending more time bouldering and others focusing on \textit{lead climbing}. The climbing abilities of the participants, in terms of their maximum redpoint grade, ranged from 6a to 7b for \textit{top rope climbing} and 6a to 7a for \textit{lead climbing} which are classified as intermediate to advanced \cite{draper2015IRCRA}, indicating that the cohort had a broad range of skill levels. \footnote{In climbing, the term redpoint grade refers to the highest difficulty level a climber can successfully ascend after prior attempts, in this work denoted by the French grading system. The number represents the difficulty level with higher numbers and letters indicating harder routes.} Participants with more climbing experience tended to have higher redpoint grades, which aligns with typical climbing progression patterns. Fifteen out of nineteen participants reported participating in other extreme sports. 

\subsubsection{Experimental Protocol}
The experiment was a within-participants design in which each participant completed two trials during the experimental session. Participants were tasked with climbing the same route using both \textit{lead} and \textit{top rope} ascent styles. This was performed in a pseudorandomized order so that half of the participants completed the top rope condition first and the other half completed the lead condition first This protocol allowed each participant to rest between trials to minimize confounding factors such as accumulated fatigue and prior route knowledge from influencing the experimental conditions. Prior to the experiment, the participants were informed about the procedure, data collection methods, and their rights regarding personal information. The collection of personal data was approved by Sikt - Norwegian Agency for Shared Services in Education and Research, reference number 727897. 
The participants completed a 5-minute warm-up routine before the experimental protocol. The routine included specific exercises designed to target various muscle groups involved in climbing and to minimize the risk of injury. The warm-up protocol is described in the Appendix \ref{sec:warm-up-protocol}. Participants were fitted with wearable physiological sensors. The trials were video-recorded and began and ended with a sensor synchronization protocol \ref{sec:synchronization}.

Before each ascent trial, a safety check was conducted in collaboration with the experimenter. During the \textit{lead climbing} trials, the climbers clipped the rope into quickdraws along the route. For the \textit{top rope} climbing trials, the rope was already secured to the top anchor of the route and participants were required to unclip the rope from the quickdraws as they ascended to match the challenge of clipping the rope in the \textit{lead} ascent style. After the climb, the participants were lowered, and the sensor synchronization protocol was repeated.
At the end of the experimental session, participants completed the post-experiment questionnaire where they reported their demographic information, previous climbing experience, and a short version of the State-Trait Anxiety Inventory, adapted for this study to assess climbing specific contexts. The adjusted STAI employed in the experiment is attached in the Appendix \ref{sec:stat_analysis}. The STAI is a well-established tool for measuring trait anxiety \cite{zsido2020development}. Elevated scores on the \gls{stai} correlate with higher vulnerability to anxiety disorders and greater emotional reactivity to stress \cite{spielberger1971state, chambers2004relationship}. 

\subsubsection{Self-report Measures}\label{sec:perception-measures}
After each trial, participants completed the post-climb questionnaire as they reviewed the video recording of their ascent. The video was paused after each quarter of the route as defined by equidistant points on the wall. Each quarter had four corresponding questions, attached in the Appendix \ref{sec:anxiety_thermometer}. With assistance from the anxiety thermometer which has ten response options \cite{houtman1989anxiety}, participants reported their anxiety level, fear of falling due to fatigue, and fear of heights.Participants also reported their feeling of fatigue in the forearms (i.e., 'pump') on a 10-point scale with response poles of 'Not pumped at all' to 'Extremely pumped'.   

\subsubsection{Physiological Measures}
The BioPoint\texttrademark{} biosensor was used to collect physiological data during the climbing trials. This compact wearable device, roughly the size of a smartwatch, is equipped with several high-quality sensors, including \gls{ecg}, \gls{ppg}, skin temperature, a six-axis \gls{imu}, and \gls{emg}. The device has 4 GB of internal memory, making it well-suited for continuous data collection during the climbing experiment \cite{SiFiLabs2024}. 
The climbers wore the BioPoint on their right forearm to measure their limb movement. We positioned the sensor on the participants’ \gls{fds}, approximately $\frac{3}{4}$ of the way from the elbow to the wrist, toward the ulnar side. We confirmed placement by palpating the muscle while the participants flexed their fingers and applied pressure with the palm of their hand, which helped to accurately determine the muscle position for sensor placement \cite{boeker2024predictive}.

\subsubsection{Data Processing and Analysis}\label{sec:feature_generation}
The data collected from  the\gls{emg}, \gls{imu}, and \gls{ecg} signals were pre-processed to reduce noise and extract meaningful features. The \gls{emg} signal was sampled at a frequency of 2000 Hz. A 50 Hz notch filter was applied to remove electrical interference, followed by a bandpass filter with a range of 20 to 450 Hz and an order of 4 \cite{li2011effects, merletti2020tutorial}. For the \gls{imu}, which was sampled at 50 Hz, a bandpass filter with a bandwidth of 0.1 to 25 Hz was applied \cite{boeker2024predictive}. The \gls{ecg} signal was processed with a bandpass filter from 1 Hz to 30 Hz at a sampling rate of 500 Hz.
To derive meaningful features from these signals, we used a rolling window approach. For the \gls{emg} signal, the sliding window was configured to a size of 1000 time points, with an overlap of 500 time points. Similarly, the \gls{ecg} signal used a sliding window of 250 time points, with an overlap of 125 time points. Each window, sliding across the signal, was used to calculate several time-dependent features. These included the \gls{mnf}, and \gls{iemg} for \gls{emg}, which are commonly used to measure muscle fatigue and muscle activity. The \gls{mnf} is a well-established measure of muscle fatigue \cite{roy1990fatigue, de1984myoelectrical}, and \gls{iemg} measures muscle activation based on the neuronal input to the muscle \cite{viitasalo1977signal}.
For the \gls{ecg} signal, we extracted both the heart rate and heart rate variability. Heart rate was calculated based on the intervals between successive R-peaks detected in the ECG waveform, while heart rate variability was assessed using the \gls{rmssd} between heartbeats \cite{makivic2013heart}.
For the \gls{imu}, we specifically analyzed the wrist's acceleration in the horizontal x-direction. The acceleration was averaged over defined time windows to reduce noise and improve the reliability of the measurements.

\subsection{Modeling}
The methodology section is divided into two parts. First, we introduce the adaption of random effects in deep learning and describe how they are integrated into the architecture and what changes were made to the loss function to accommodate them. We conclude with a description of the model evaluation methods used in this study.

\subsubsection{Random effects in deep learning}
One of the advantages of machine learning is its minimal requirement for statistical assumptions. However, every model assumes that the data points are \gls{iid}, which means that the data points are drawn independently from the same distribution \cite{bishop2006pattern}. This assumption is violated when data are clustered or temporally correlated. A violation of this assumption can lead to biased estimates and false inferences \cite{kilian2023mixed}.

Fortunately, machine learning offers several mechanisms to handle such scenarios. For instance, network architectures such as \gls{rnn} or \gls{gru} are explicitly designed to model serial dependencies between observations. These models capture the temporal structure of the data, ensuring that predictions consider the order and dependencies between data points, thus handling violations of the i.i.d. assumption in time-series data.

For clustered data, in which observations within clusters are correlated, a common approach in machine learning is to encode clusters as categorical features.
These categorical features can be transformed using methods such as one-hot coding or entity embedding \cite{hancock2020survey, geron2022hands}. 
Unlike embeddings, which learn dense representations for each cluster, the random effects model clusters membership as a random variable, allowing the model to capture within-cluster correlation. This approach is particularly useful for hierarchical data or repeated measures data, where observations within the same cluster are not independent.   

The concept of random effects allows us to account for the correlated data $X$ when modeling the dependent variable $Y$. Accordingly, $\boldsymbol{y}_j$ is assumed to be a multivariate normal variable $\boldsymbol{y}_j \sim N(\boldsymbol{\mu}_j, V(\psi))$, where the mean $\boldsymbol{\mu}_j$ has a size $(n_j \times 1)$ and a covariance matrix $V(\theta)$ has a size $(n_j \times n_j)$. $n_j$ is the number of observations per cluster $j$ \cite{laird1982random}. $J$ is the number of clusters. 
We assume that $b_j$ is drawn from $b$ a multivariate normal distribution $N(0, \boldsymbol{D})$. $\boldsymbol{D}$ is $(j \times j)$ positive semi-definite covariance matrix \cite{simchoni2023integrating}. 
We can translate linear mixed effects models into a non-linear setting, by allowing fixed and random effects to be non-linear. 
In doing so, we followed the approach of Simchoni et al. \cite{simchoni2021using}. ¨
In Equation \ref{eq:nonlinear_mixed_effects}, we generalize the equations to represent the data without explicitly indexing the clusters $j$. This means that $\boldsymbol{X}$ refers to the overall design matrix for fixed effects, $\boldsymbol{Z}$ represents the design matrix for random effects across all observations, and $\boldsymbol{b}$ and $\epsilon$ denote the vectors of random effects and noise, respectively.

\begin{equation}\label{eq:nonlinear_mixed_effects}
    \boldsymbol{y} = f(\boldsymbol{X}, \boldsymbol{\theta}) + \boldsymbol{Z} \boldsymbol{b} + \epsilon
\end{equation}
In Equation \ref{eq:nonlinear_mixed_effects}, $f(\boldsymbol{X}, \boldsymbol{\theta})$ and $g(\boldsymbol{Z})$ are potentially complex functions that can be substituted with a neural network architecture. For simplicity and interpretability, we chose $g$ as the identity function, so that  $g(\boldsymbol{Z}) = \boldsymbol{Z}$. 
When integrating random effects, we need to optimize the neural network parameters $\boldsymbol{\theta}$ and the random effect vector $b$. Since $b \sim N(0,D)$, we optimize for the variance-covariance parameters in $D$. For example, in the case of a random intercept, we optimize for $\sigma_b I$. Instead of minimizing a loss function such as \gls{mse}, we minimize the negative log likelihood given in Equation \ref{eq:NLL}, where $V(\psi) = ZD(\psi)Z' + \sigma_e^2 I$ \cite{simchoni2021using}. 
\begin{equation}\label{eq:NLL}
    \begin{split}
        NLL(f ,\theta, D| y) = & \ \frac{1}{2} ( y - f(X,\theta))'V(\psi)^{-1}(y-f(X, \theta)) \\
        & + \frac{1}{2} \log |V(\psi)| + \frac{n}{2} \log 2 \pi
    \end{split}
\end{equation}
The covariance of the random effects takes the general structure described in Equation \ref{eq:RE_cov_matrix}. 
\begin{equation}\label{eq:RE_cov_matrix}
    D(\psi) = \begin{pmatrix}
\sigma_{b_1}^2 I_j & \cdots & \rho_{1,J-1}\sigma_{b_1}\sigma_{b_{J-1}} I_j\\
\rho_{1,2}\sigma_{b_1}\sigma_{b_2} I_j & \sigma_{b_2}^2 I_j  & \rho_{2,J-1}\sigma_{b_2}\sigma_{b_{J-1}} I_j\\
\cdots & \cdots & \cdots \\
\rho_{1,J-1}\sigma_{b_1}\sigma_{b_{J-1}}I_j & \cdots & \sigma_{b_{J-1}}^2 I_j
\end{pmatrix}
\end{equation}
Depending on the covariance structure of the random effects in $V(\psi)$ the inversion can be computationally expensive. However, for the common random intercept and random slope models, we can decompose $V(\psi)$ into the sum of block diagonal matrices. This decomposition simplifies the inversion process. 

In the random intercept model, the covariance matrix becomes the diagonal matrix $\sigma_{b_1}^2 I_j$, whose inversion is the element-wise inversion of the diagonal elements, resulting in $\frac{1}{\sigma_{b_1}^2} I_j$. In the case of the random slopes model, $D(\psi)$ is given by the matrix in Equation \ref{eq:random_slopes_cov}, which can be inverted block-wise.
\begin{equation}\label{eq:random_slopes_cov}
\begin{pmatrix}
\sigma_{b_0}^2 & \rho_{0,1} \sigma_{b_0} \sigma_{b_1} \\
\rho_{0,1} \sigma_{b_0} \sigma_{b_1} & \sigma_{b_1}^2
\end{pmatrix}
\end{equation}

\subsubsection{Deep learning model and training specifications}
The machine learning experiments evaluate the performance of a \gls{gru} to compare benchmark models for predicting the sum of the perceived fear scores using time series data and demographic data. We included \gls{mnf}, \gls{hr} and acceleration of the arm as time series features. Age and gender were included as demographic variables.

Recurrent architectures such as \gls{gru} are well suited to model temporal dependencies within physiological signals that simpler models potentially cannot capture. We chose \gls{gru} because it has fewer parameters than the well-known \gls{lstm}. However, both architectures share the important additive update mechanism, which helps them remember features for long sequences and creates shortcut paths that mitigate vanishing gradient problems \cite{chung2014empirical}.

We evaluated the performance of the \gls{gru} and compared it to a linear model and a \gls{mlp}. Each model processed fixed time windows of 30 sequential measurements. Our architectures were designed with a multi-branch approach. One branch processes time series data while another handles participant metadata, before merging for final prediction.
All physiological signals were collected synchronously from an integrated sensor unit. After preprocessing and feature extraction as described in Section \ref{sec:feature_generation}, we created fixed-length windows of 30 sequential measurements for model input. Each input window contained the standardized features from \gls{emg}, \gls{ecg}, and \gls{imu} sensors, creating a multivariate time series representation of dimensions sequence length and number of features. These windows were generated using a sliding window approach with no overlap between consecutive windows. For each window, we derived a single target value by taking the sum of the perceived fear scores. 

The linear model directly averages time series features and applies a single linear transformation. The \gls{mlp} processes time series data through a dense layer with ReLU activation, averaging across time points, before merging with metadata features processed through a separate dense layer. The merged representation passes through another dense layer with ReLU activation and dropout of $50\%$.
The GRU model processes time series data through a single GRU layer. This captures the entire sequence information in a compact representation. The model then combines this with participant metadata for the final prediction. This concatenated representation is then processed through a linear layer with ReLU activation and subsequently passed through a dense layer that also employs ReLU activation and a dropout rate of $50\%$ before making the final prediction.

In addition, each model was compared to its counterpart with random effects that account for subject-specific variability. These models include a custom RandomEffectLayer that adds participant-specific intercepts to the outputs, while standard models focused exclusively on fixed effects.

Training was performed using Adam optimization with early stopping to prevent overfitting. The learning rate was $lr = 0.0001$ with an $L2$ regularization of $\lambda = 0.0001$. All deep learning architectures include a dropout rate of $ 50\%$ for overfitting prevention. We used \gls{mse} loss for standard models, while random effect models were trained with negative log-likelihood loss.

We split the participant's ascent trials in a $70\%$ training, $15\%$ validation, and $15\%$ test distribution. Our approach acknowledges that participants completed multiple climbing routes in different styles. Rather than separating participants entirely between sets, we randomly added one ascent style for every climber in the training set. Then we assigned their remaining climbs and split these between training, validation, and test sets to achieve our $70\%$, $15\%$, $15\%$ split.
Each experiment was repeated ten times with different random weight initializations, with averages and standard deviations calculated across runs. The models were trained using a batch size of 64 for up to 100 epochs.

\section{Results}
The result section is structured into three parts. In Section \ref{sec:stat_analysis}, we present a statistical analysis using a linear mixed-effects model that provides interpretable insight into the relations between physiological signals, the ascent style and height, and self-reported fear. We then explore the application of deep learning techniques and their integration with random effects to improve the model performance in Section \ref{sec:dl-compare}. Finally, we evaluated these deep learning models using advanced metrics that go beyond traditional regression analysis and provide a more sophisticated understanding of the data.

\subsection{Statistical Analysis}\label{sec:stat_analysis}
\begin{figure}[H]
    \centering
    \includegraphics[width=1\linewidth]{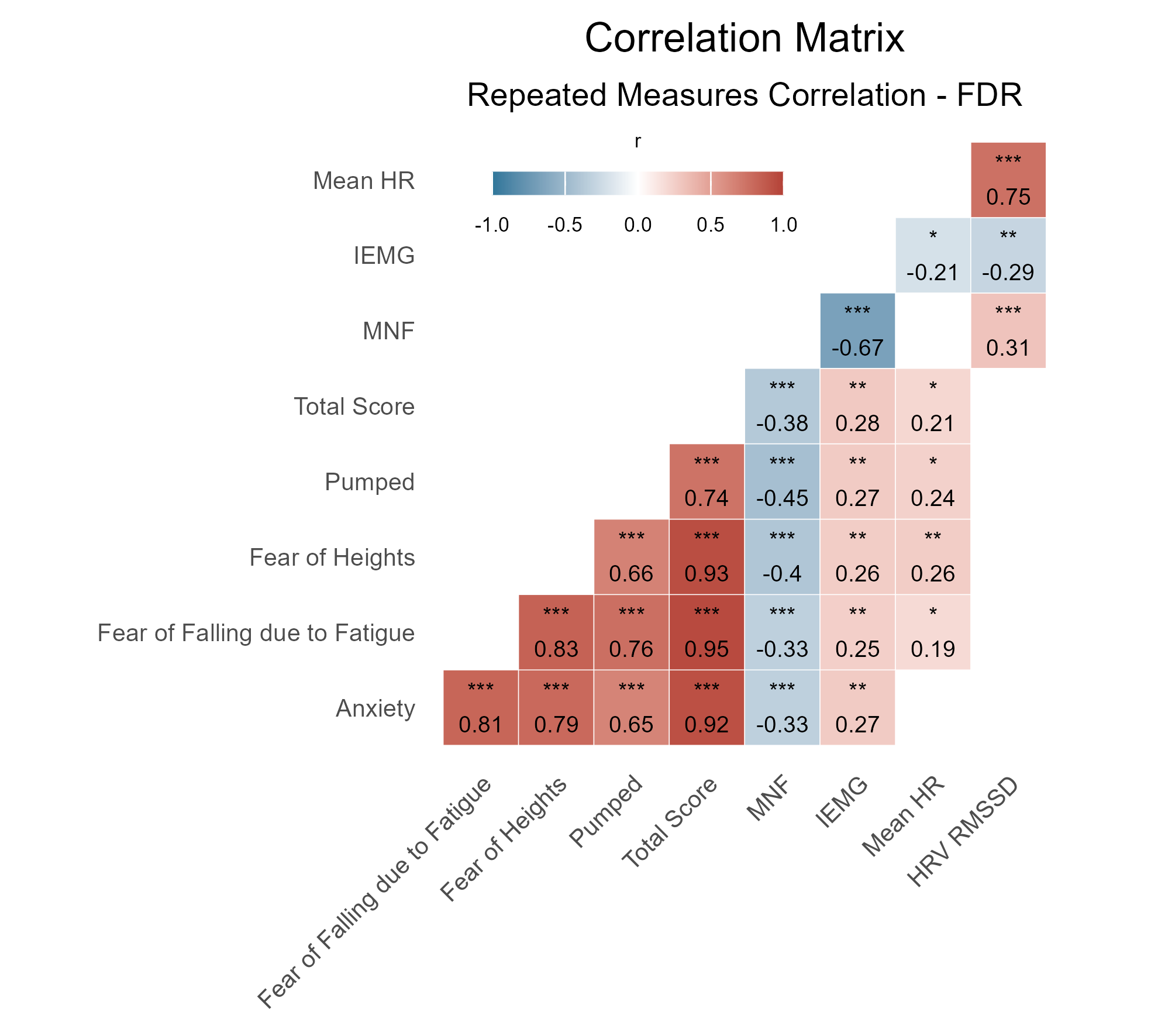}
    \caption{The correlation matrix illustrates pairwise repeated-measures correlation coefficients and false discovery rate corrected p-values between the subjective and physiological variables. Each cell in the heatmap represents the correlation coefficient and significance of that specific pair. Only significant correlations are visualized. * p < .05, ** p < .01, * p < .001..\label{fig:correlation plot}}
\end{figure}

\begin{figure*}
        \centering
        \begin{subfigure}[b]{0.475\textwidth}
            \centering
            \includegraphics[width=\textwidth]{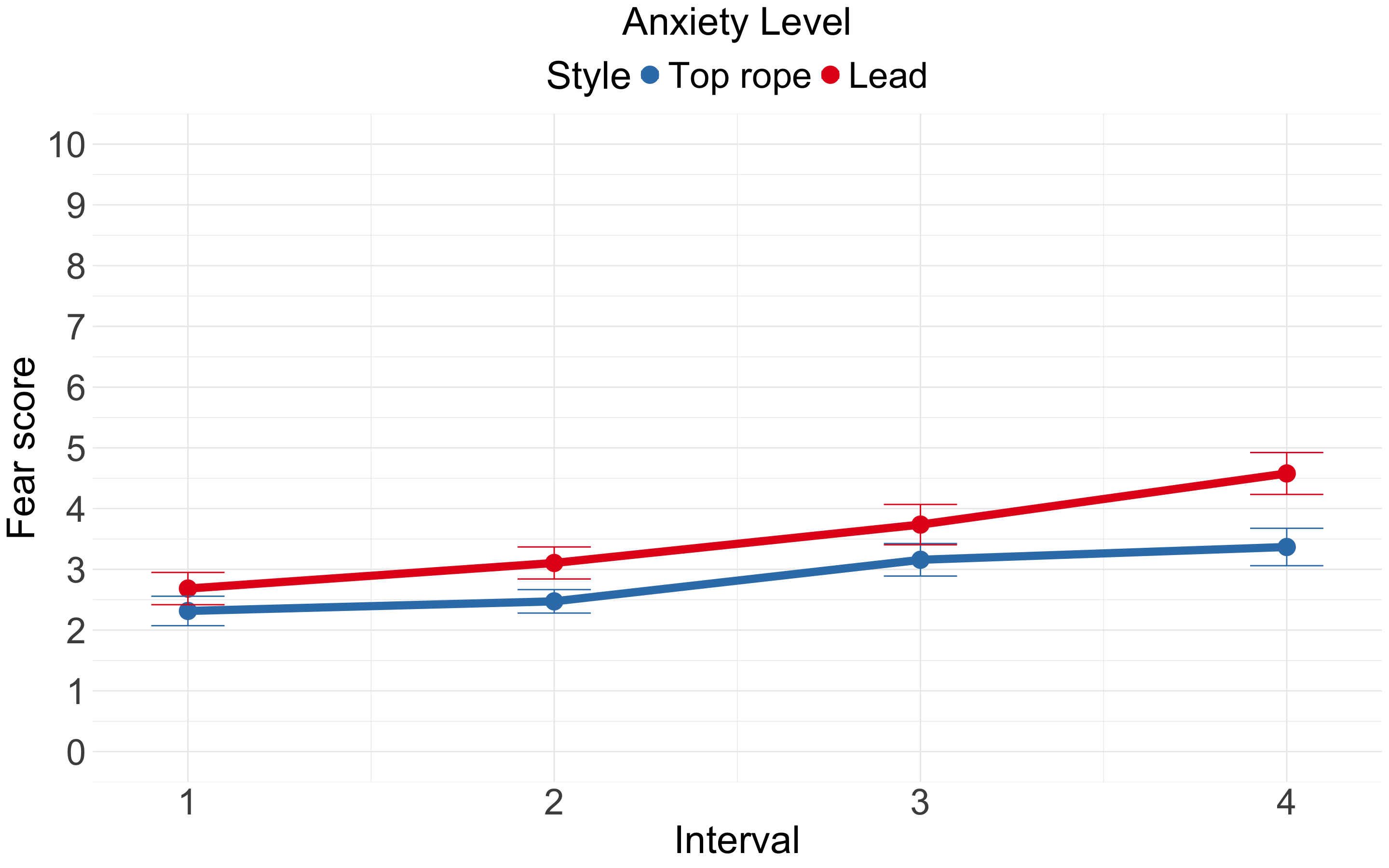}
            \label{fig:general_fear_level}
        \end{subfigure}
        \hfill
        \begin{subfigure}[b]{0.475\textwidth}  
            \centering 
            \includegraphics[width=\textwidth]{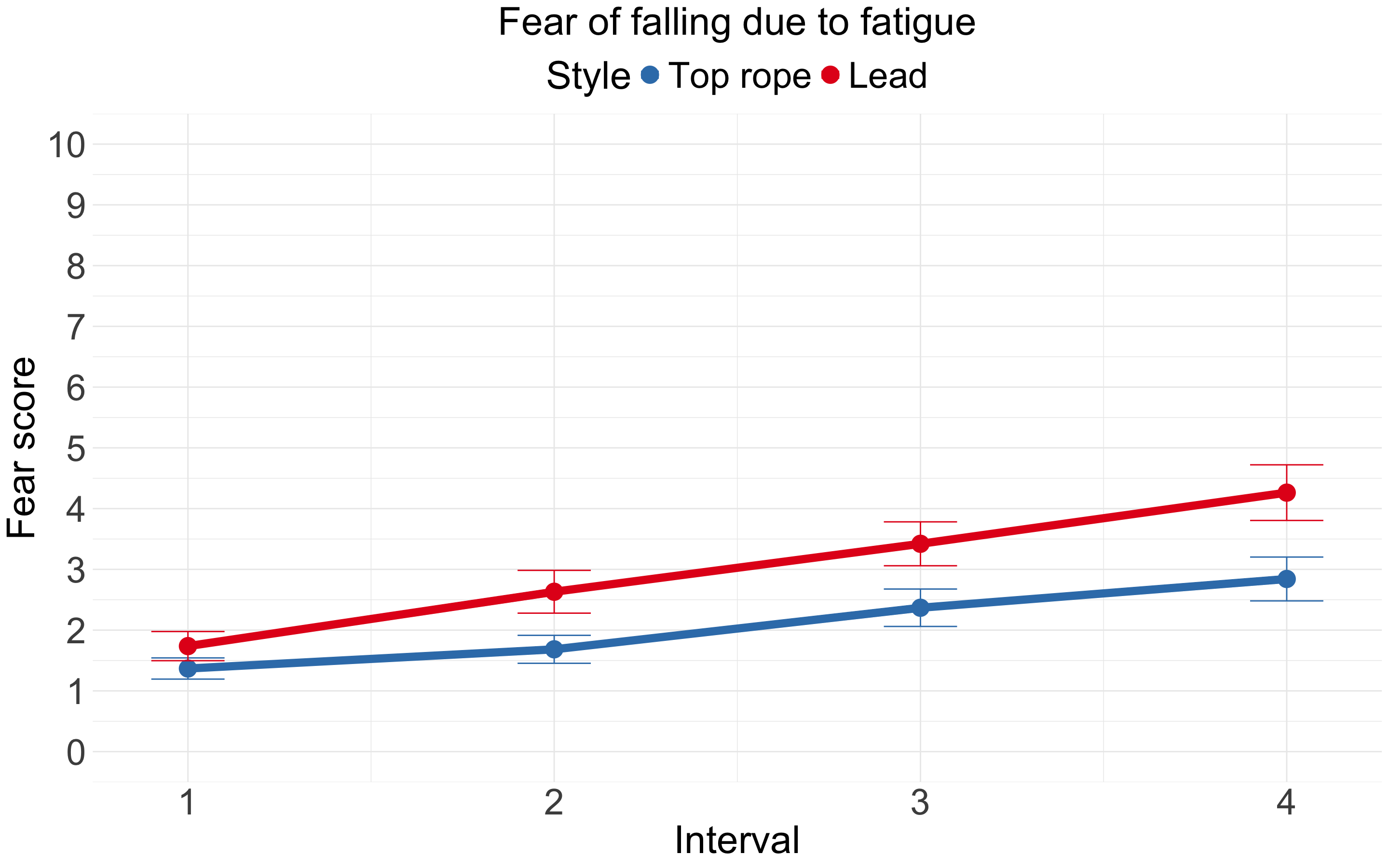}
            \label{fig:fear_of_falling}
        \end{subfigure}
        \vskip\baselineskip
        \begin{subfigure}[b]{0.475\textwidth}   
            \centering 
            \includegraphics[width=\textwidth]{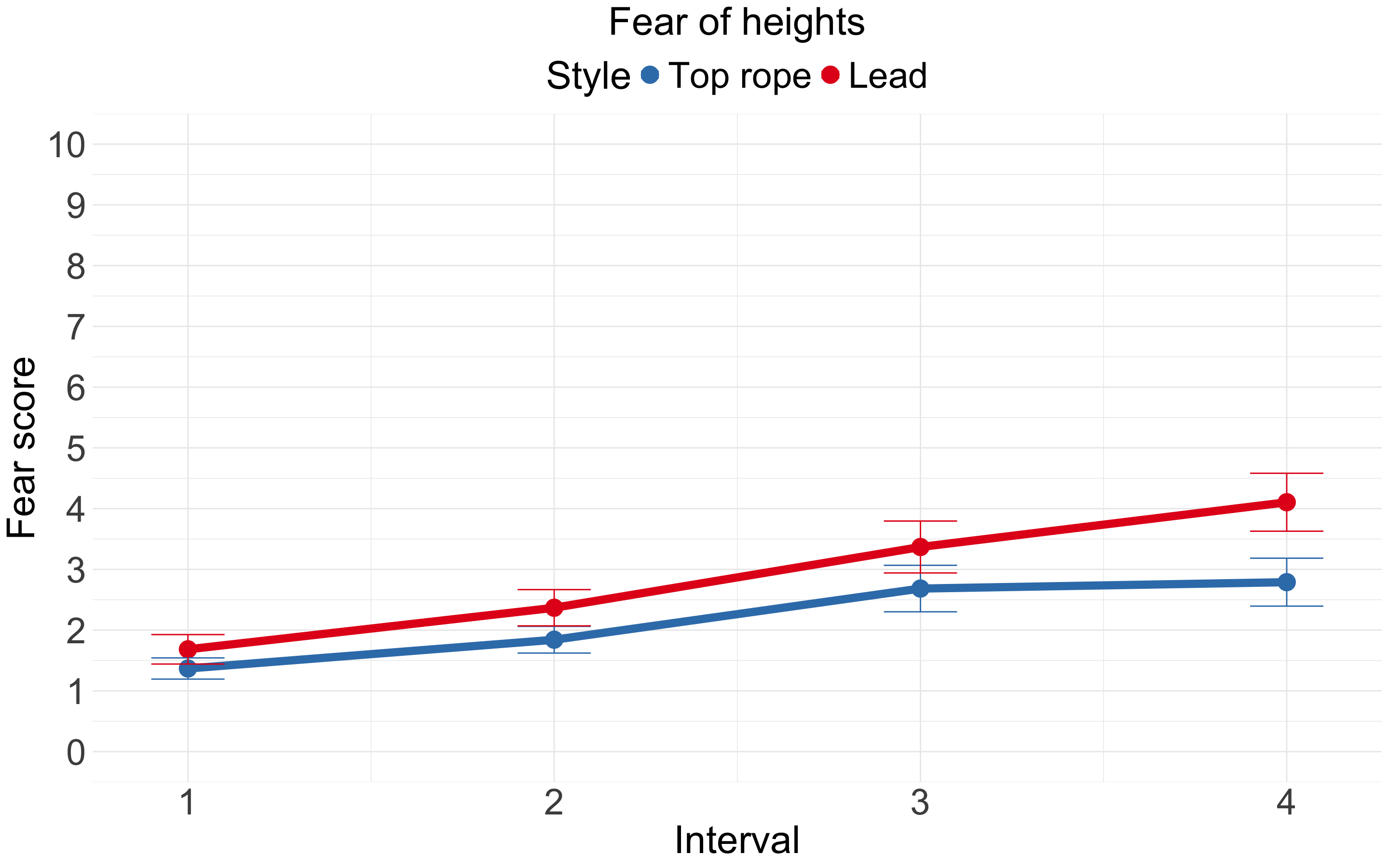}
            \label{fig:fear_of_heights}
        \end{subfigure}
        \hfill
        \begin{subfigure}[b]{0.475\textwidth}   
            \centering 
            \includegraphics[width=\textwidth]{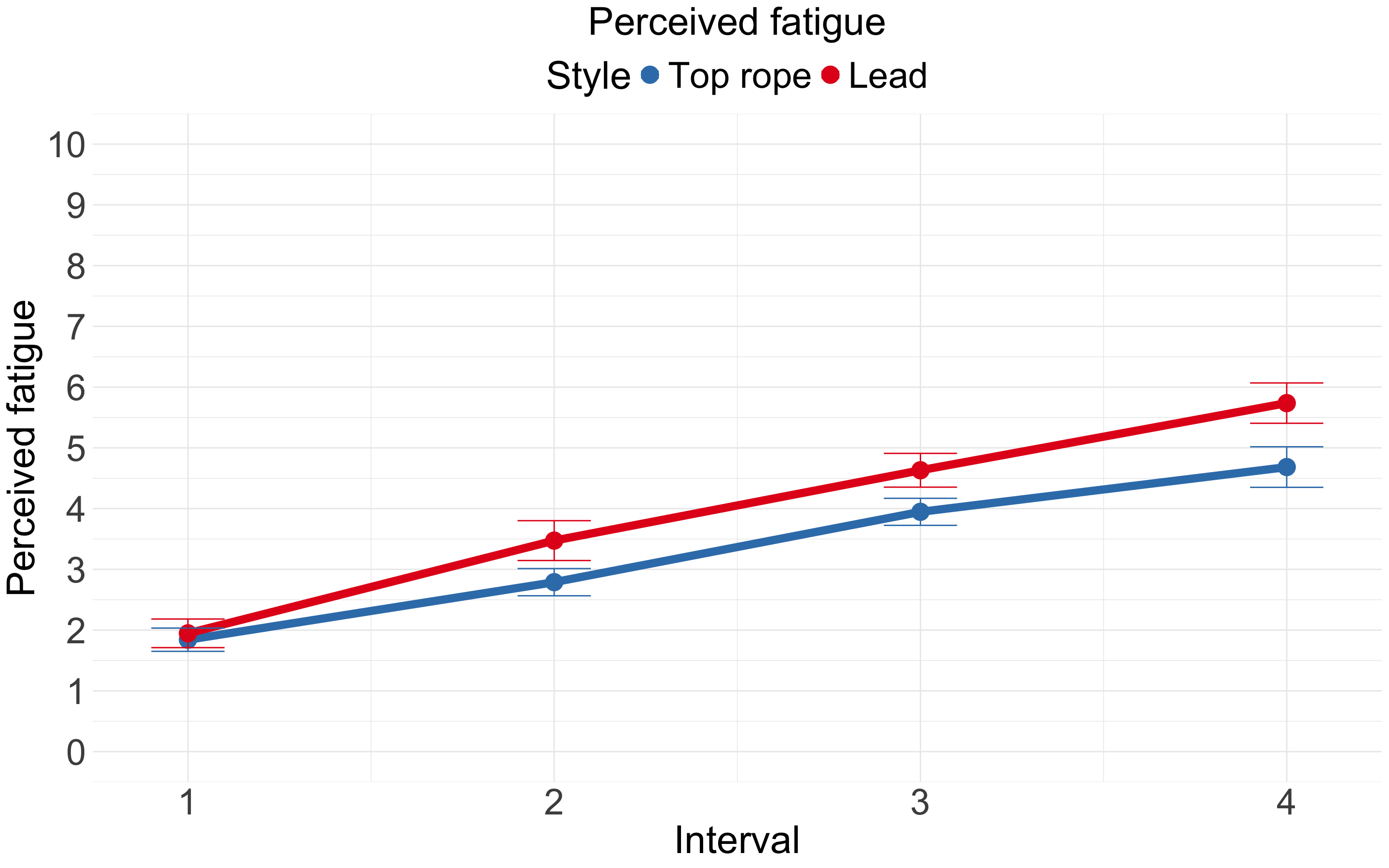}
            \label{fig:subjective_fatigue}
        \end{subfigure}
        \caption[]
        {\small The figure compares the four subjective scores measured at each interval and for \textit{lead} and \textit{top rope climbing}, displayed as line plots with standard error bars. We observe that participant consistently reported higher scores during \textit{lead climbing} than \textit{top rope climbing}. } 
        \label{fig:mean and std of nets}
    \end{figure*}

Before conducting, statistical modeling, we describe the collected data to gain a general understanding of the characteristics and relations. Figure \ref{fig:mean and std of nets} shows the results of the four self-report measures described in Section \ref{sec:perception-measures} over the intervals with their standard errors. The self-report measures increase with increased height and that the trajectories are steeper in \textit{lead} than \textit{top rope climbing}. The figure also indicates a similar linear increase for all the four perceived measurements.
Figure \ref{fig:correlation plot} illustrates the pairwise repeated measures correlation matrix as a heat map. It displays the correlation coefficients along with \textit{p}-values corrected for multiple comparisons using the false discovery rate approach. Participants 3 and 6 were removed from the correlation because of noisy \gls{ecg} signals. First, we observed strong correlations among the perceived measurements, especially fear-related scores. The correlation matrix also revealed a positive linear relationship between increasing perceived fear and muscle activity measured by \gls{iemg}. A decreasing \gls{mnf} as a measure of increasing muscle fatigue, is also correlated with perceived fear and perceived fatigue (i.e., "Pumped"). These findings suggest that muscle activity and fatigue have a linear relationship with perceived fear. 

As described in Section \ref{sec:perception-measures}, the questionnaire measurement response scales ranged from one to ten. Given the high correlation among the three fear scores—general anxiety, fear of falling due to fatigue, and fear of heights—we explored the possibility of aggregating them into a single score. Aggregating these scores provides a more reliable representation of the underlying construct and approximates a continuous scale, compared to treating each score individually. Therefore, we conducted both Cronbach's alpha and principal component analysis to assess whether the scores measured the same underlying construct \cite{tavakol2011making}. The analysis revealed good internal consistency with a Cronbach's alpha of 0.92. Moreover, principal component analysis confirmed that the scores loaded onto a single component, explaining $87\%$ of the variance. Building on this analysis, we aggregated these scores to make a total fear score. 

We used linear mixed-effects models to analyze the effects of climbing style, interval distance, climbing experience in years, muscle activity and fatigue on the total score of perceived fear.
We fitted two different mixed-effects models with the total score as the outcome measure and \gls{mnf} and \gls{iemg} as predictors. 
Both models included \textit{style} (lead = 1, top rope = 0) and \textit{interval} (four levels) as categorical fixed effects. Additionally, we included an interaction term between the muscle activity predictors, \gls{iemg} or \gls{mnf}, and \textit{Style} to assess whether the relationship between muscle activity and the total score varies depending on the climbing style. A random intercept was included for each participant to account for individual differences.
The intercept in both models represents the predicted \textit{total score} for \textit{top rope climbing} in the first interval when \gls{iemg} or \gls{mnf} is equal to zero. The results are summarized in Table \ref{tab:regression_summary}. Participant 20 was excluded from the linear mixed effects model due to faulty \gls{emg} signal measurement.

\begin{figure}
    \centering
    \includegraphics[width=1\linewidth]{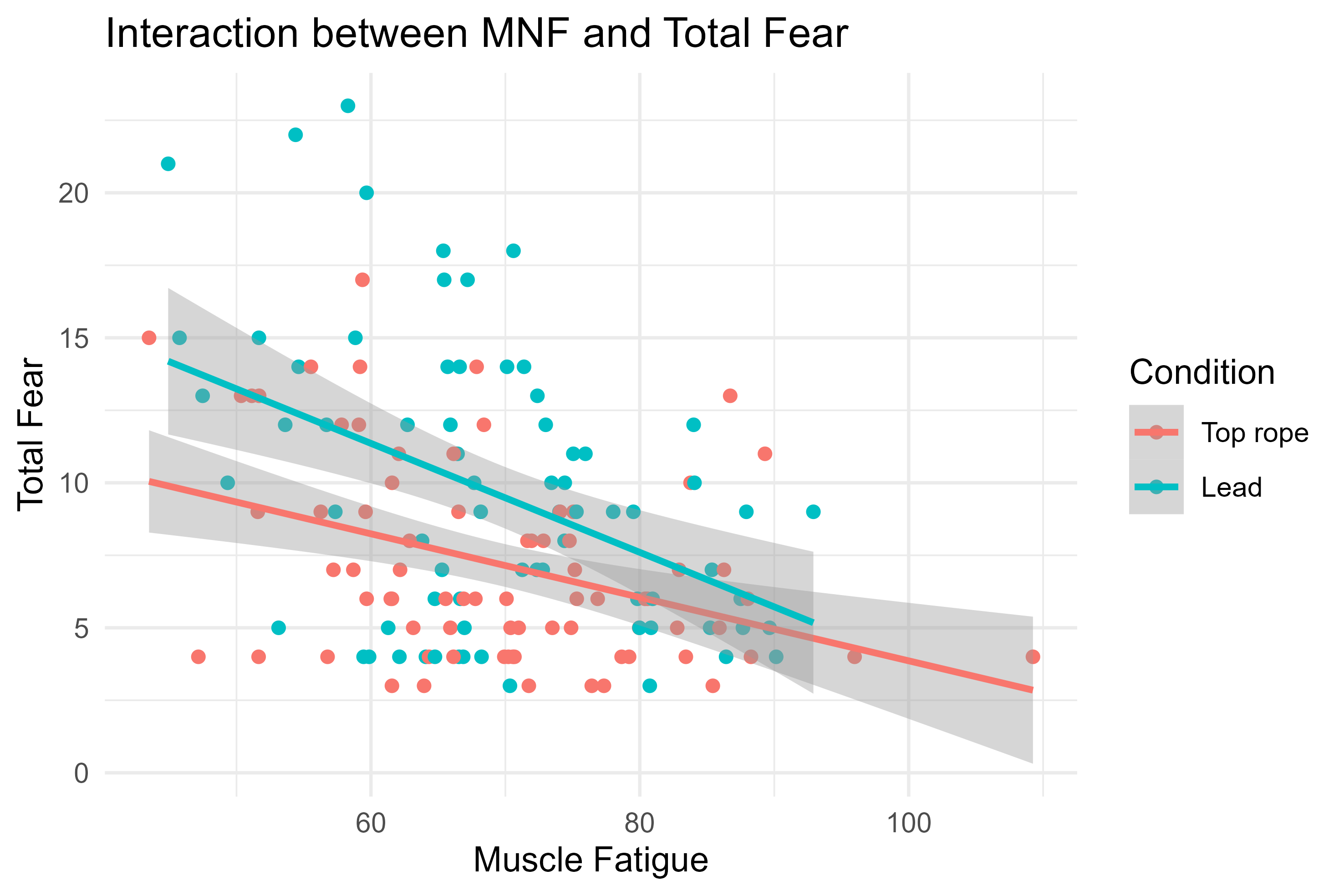}
    \caption{The relationship between \gls{mnf} and self-reported fear, conditioned by climbing ascent style. The figure illustrates a steeper negative slope for lead climbing compared to top rope climbing, indicating a stronger association between \gls{mnf} and fear in lead climbing.\label{fig:interaction-term-mnf-total-score}}
\end{figure}

For the \gls{mnf} model, climbing style significantly impacted self-reported fear, with higher levels reported for \textit{lead climbing} (\(\text{estimate} = 13.650\), \(p < .001\)) compared to \textit{top rope climbing} (\(\text{intercept} = 5.236 \), \(p = .043\)). Later intervals (3 and 4) exhibited a significant positive effect on self-reported fear, with Interval 3 (\(\text{Estimate} = 3.481\), \(p < .001\)) and Interval 4 (\(\text{Estimate} = 3.516\), \(p < .001\)) showing increased self-reported fear relative to Interval 1. This indicates that perceived fear intensifies as the climber ascends, which is possibly related to increasing height or fatigue. The significant interaction term between \gls{mnf} and \textit{lead climbing} (\(\text{estimate} = -0.089\), \(p = .019\)) suggests that muscle fatigue was more strongly associated with self-reported fear during \textit{lead climbing} than \textit{top rope climbing}. This relation is visualized in Figure \ref{fig:interaction-term-mnf-total-score}, which shows the relation between \gls{mnf} and self-reported fear. 

\begin{figure}
    \centering
    \includegraphics[width=1\linewidth]{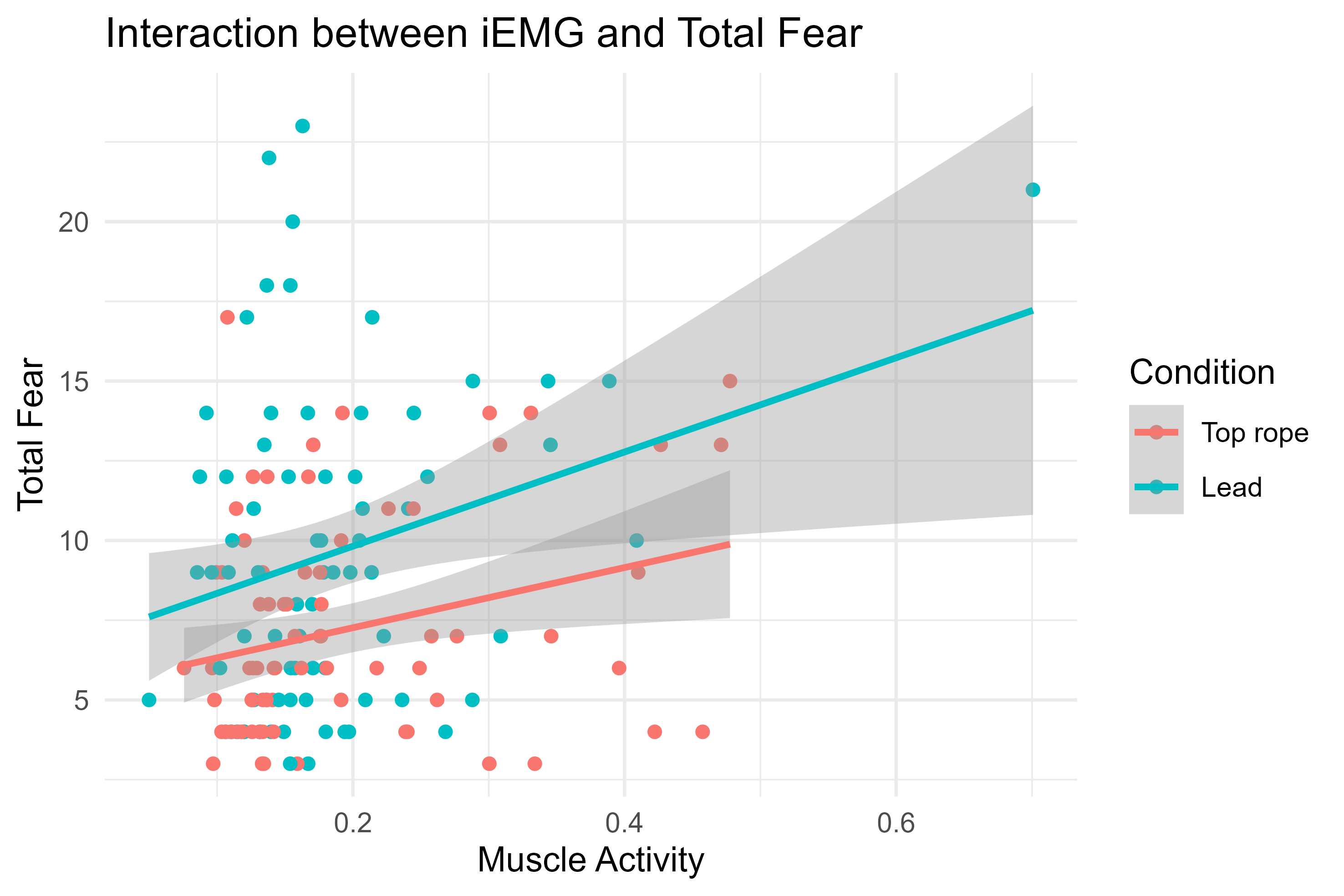}
    \caption{The figure shows the relationship between \gls{iemg} and self-reported fear across two climbing styles. Although lead climbing shows a slightly steeper increase in self-reported fear with \gls{iemg} compared to top rope climbing, the interaction term was not statistically significant. \label{fig:interaction-term-iemg-total-score}}
\end{figure}

For the \gls{iemg} model, climbing style had a significant effect, with higher perceived fear of \textit{lead climbing} (\(\text{estimate} = 7.229\), \(p < .001\)) compared to \textit{top rope climbing} (\(\text{intercept} = 4.969\), \(p < .001\)). Later intervals (3 and 4) also showed significant increases in self-reported fear, with Interval 3 (\(\text{Estimate} = 3.716\), \(p < .001\)) and Interval 4 (\(\text{Estimate} = 4.693\), \(p < .001\)) reflecting higher perceived fear as the climb progressed. The main effect of \gls{iemg} (\(p = .110\)) and the interaction between \gls{iemg} and climbing style (\(p = .512\)) were not statistically significant, indicating that muscle activity as measured by \gls{iemg} does not explain self-reported fear, nor does its relationship with fear differ by climbing style. Similar to Figure \ref{fig:interaction-term-mnf-total-score}, Figure \ref{fig:interaction-term-iemg-total-score} visualizes the interaction between the relation between \gls{iemg} and self-reported fear. 

\begin{table}[h!]
\centering
\caption{Summary of linear mixed model analyses of the effect of muscle activity on fear. The table shows the estimates, standard errors, and p-values for the regression models. Significant parameters (\(\alpha < .05\)) are annotated with a single asterisk (*).}
\label{tab:regression_summary}
\begin{tabular}{@{}lllll@{}}
\toprule
\textbf{Model}       & \textbf{Fixed Effect} & \textbf{Estimate} & \textbf{Std. Error} & \textbf{p-value} \\
\midrule
\multirow{8}{*}{MNF} 
                     & MNF                  & -0.0008            & 0.033               & .980            \\
                     & Intercept   &  5.236             & 2.595               & .046 *          \\
                     & Lead               & 13.650             & 2.822               & <.001 *       \\
                     & Interval2            &  1.054             & 0.620               & .092           \\
                     & Interval3            &  3.516             & 0.649               & <.001 *       \\
                     & Interval4            &  4.628             & 0.686               & <.001 *       \\
                     & experience           & -0.074             & 0.092               & .434            \\
                     & MNF:style1           & -0.089             & 0.037               & .019 *          \\
\midrule
\multirow{8}{*}{IEMG} 
                     & IEMG                 &  0.636             & 0.380               & .097           \\
                     & Intercept   &  4.969             & 0.944               & <.001 *       \\
                     & Lead               &  7.229             & 0.953               & <.001 *       \\
                     & Interval2            &  1.161             & 0.604               & .057           \\
                     & Interval3            &  3.716             & 0.605               & <.001 *       \\
                     & Interval4            &  4.693             & 0.631               & <.001 *       \\
                     & experience           & -0.063             & 0.088               & .485            \\
                     & IEMG:style1          &  0.283             & 0.430               & .512            \\
\bottomrule
\end{tabular}
\end{table}

\subsection{Model Comparison}\label{sec:dl-compare}
The results from the linear mixed models showed a significant relationship between perceived fear, muscle activity, and muscle fatigue. However, we also investigated the potential of deep learning models and how personalized deep learning models performed when random effects are considered. 
The performance of all models, both with and without random effects, was compared using \gls{mse}, \gls{mae}, and \gls{rmse}. The results are summarized in Table \ref{tab:Metric_Results}. The values are given as mean $\pm$ standard deviation across ten different random initializations to provide an overview of the model performance and variability between experiments. The models were evaluated on the completely held-out test set (15\% of the data), unlike the analysis in Section \ref{sec:stat_analysis}, which was performed on the entire dataset. 

The training and validation loss curves for the deep learning models across training epochs are added to the Appendix \ref{sec:training_validation_loss}. The consistent decrease in validation loss without significant divergence from training loss indicates proper convergence without overfitting.

The \gls{gru} with random effects achieved the lowest \gls{rmse} ($2.49 \pm 0.27$) as well as the lowest \gls{mae} ($1.42 \pm 0.18$) and \gls{mse} ($6.25 \pm 1.39$), indicating superior predictive performance. All random effects models consistently outperformed their standard counterparts across all three metrics, demonstrating the importance of accounting for participant-specific variability in physiological response modeling.

To contextualize these error metrics within the scale of our target variable, we included summary statistics of the fear scores alongside performance metrics in Table \ref{tab:summary_and_context}. The target value has a standard deviation of 4.37 and interquartile range of 6 (Q3-Q1). The random effects \gls{gru} model achieves a \gls{rmse}/SD ratio of 0.57 indicates that prediction errors represent approximately $57\%$ of the natural variability in the data. Similarly, the \gls{rmse}/IQR ratio of 0.42 demonstrates that errors are less than half the interquartile range. These normalized metrics show that our best model achieves meaningful predictive accuracy despite the inherent variability in fear responses during climbing. The consistently superior performance of models incorporating random effects highlights the importance of accounting for participant-specific variability when modeling physiological responses to climbing-induced fear, particularly with limited sample sizes. This approach effectively captures both shared patterns across participants and individual differences in fear response profiles.
\begin{table}[htbp]
\centering
\begin{tabular}{@{}lccccccc@{}}
\hline
Metric & GRUNN & GRUNNRE & Linear & LinearRE & NN & NNRE \\
\hline
\text{MSE} & $11.19 \pm 0.99$ & $6.25 \pm 1.39$ & $27.04 \pm 0.46$ & $19.24 \pm 4.08$ & $11.52 \pm 1.07$ & $8.96 \pm 2.42$ \\$
$\text{MAE} & $1.91 \pm 0.11$ & $1.42 \pm 0.18$ & $4.23 \pm 0.01$ & $2.94 \pm 0.4$ & $2.09 \pm 0.08$ & $1.78 \pm 0.21$ \\$
$\text{RMSE} & $3.34 \pm 0.15$ & $2.49 \pm 0.27$ & $5.2 \pm 0.04$ & $4.37 \pm 0.47$ & $3.39 \pm 0.16$ & $2.97 \pm 0.39$ \\
\hline
\end{tabular}
\caption{Comparison of prediction performance between random effect models and the standard models across different error metrics. The results show a consistent improvement in all three evaluation metrics for the random effects compared to the standard models. \label{tab:Metric_Results}}
\end{table}

\begin{table}[htbp]
\centering
\label{tab}
\begin{tabular}{lccccc}
\hline
Metric & GRUNNRE & NNRE & LinearRE \\
\hline
RMSE & 2.49 & 2.97 & 4.37  \\
RMSE/SD Ratio & 0.57 & 0.68 & 1.0 \\
RMSE/IQR Ratio & 0.42 & 0.50 & 0.73 \\
\hline
Median & SD & Q1 &  Q3 \\
 7 & 4.37 &  5 & 11 \\
 \hline
\end{tabular}
\caption{Model Performance Comparison for Total Fear Score Prediction. The table presents \gls{rmse} values and the \gls{rmse}/ standard deviation ratio as well as the \gls{rmse}/ interquartile range ratios) for our three random effects models, alongside summary statistics of the target variable. \label{tab:summary_and_context}}
\end{table}


\section{Discussion}
In this study, we investigated the relation between muscle activity, muscle fatigue, and self-reported fear during different climbing ascent types and heights. We measured the physiological responses and self-reported fear in 20 climbers during both \textit{top rope} and \textit{lead climbing}. Using a linear mixed-effects model, we identified a significant relation between muscle fatigue and self-reported fear. Recognizing the complexity of these psychophysiological responses, we also used deep learning techniques extended with random effects to better capture the non-linear dynamics of the interplay of these measures, following the approach proposed by Simchoni et al. \cite{simchoni2021using}. 

Notably, the results in Section \ref{sec:stat_analysis} indicate that muscle fatigue has a significant interaction effect with climbing style, suggesting that the relationship between fatigue and self-reported fear is more pronounced in \textit{lead climbing} than in \textit{top rope climbing}. This implies that climbers experienced more fear as muscle fatigue increased during \textit{lead climbing} as compared to \textit{top rope climbing}. This may be related to how falling distances are longer in lead climbing and falls are more likely if muscles are fatigued. 
Moreover, the results showed that height intervals and \textit{lead climbing} led to significantly increased self-reported fear. These results are consistent with the findings of Hodgson et al. in which different rope safety protocols influenced fear \cite{hodgson2009perceived}. However, the linear mixed effects model failed to find a significant relation between muscle activity during \textit{lead climbing} and self-reported fear. This suggests that muscle activity, as captured by \gls{iemg}, does not directly correspond to self-reported fear responses. Instead, the cumulative effect of muscle fatigue, as captured by the \gls{mnf}, better explains self-reported fear in the more demanding context of \textit{lead climbing}. 
These results contribute to the literature on intermediate climbers that shows that \textit{lead climbing} evokes more fear and activates physiological markers of the \gls{ans}, due to perceived danger and an increased risk of falling. This activation manifests as a physiological response measurable with \gls{emg} and contributes to a stronger experience of fear.

The results in Section \ref{sec:dl-compare} show that models with random effects perform better than those without random effects. We observed an improvement in most models and feature combinations. Our findings replicated the results of Simchoni et al. as we successfully showcased the integration of random effects into deep learning to improve personalized modeling \cite{simchoni2023integrating}. This advance is particularly crucial in the field of psychophysiology research, where datasets are usually characterized by numerous repeated measurements per participant but a limited number of study participants in total. Consequently, accounting for inter-individual variability is essential to ensure model robustness and prevent biased generalization. By incorporating random effects, our approach provided a more accurate representation of individual differences, thereby improving the generalizability and reliability of the models. Furthermore, our results highlight the need to develop methods that strike a balance between personalization and generalization, which is crucial for applying models in diverse populations.

The \gls{gru} architecture achieved the best results because it is inherently well suited to capture the temporal dynamics of physiological signals. Unlike \gls{mlp} or linear models, \gls{gru} are designed to retain information across time steps, allowing them to model how earlier physiological states influence later emotional responses - a key feature of fear responses in climbing. By incorporating random effects into the GRU, the model gains the ability to learn not only from general patterns in the data, but also from participant-specific trends.

Our study has limitations regarding the investigation of the relationship between muscle activity and psychological state. These limitations reveal potential future research directions for more in-depth analysis of the psychophysiology of climbing. Extending the experiment to include a real-time measurement of portable neural measurement equipment (e.g., fNIRS) during climbing could provide a deeper insight into the participant's \gls{cns} activity. Self-reports after climbing, such as perceived fear, only provide an approximation of the emotional states. They are subject to memory bias and cannot fully capture the dynamic nature of the psychological states during climbing. Consequently, it is not possible with these reports to determine exactly when fear is experienced in relation to muscle activity. Future work could aim to have climbers report their fear experience in real-time to allow for more nuanced time-series analyses. Regarding the random effects in deep learning, future work should aim to integrate random effects for classification task of neural networks. The current approach integrates the random effects into the negative log likelihood function of a regression \ref{eq:NLL}. Especially multi-class classification problems or Poisson distributed dependent variables, to improve the extent to which they can be personalized \cite{simchoni2021using, paul2011predictive, greene2001fixed}. 

\section{Code Availability}

\appendix
\section{Appendix}

\subsection{Warm-up Protocol}\label{sec:warm-up-protocol}
The warm-up lasted approximately 5 minutes and consisted of tendon gliding to mobilize the fingers and hands, fist squeezes to warm up the forearms, 20 jumping jacks to activate the full body, arm circles to mobilize the shoulders, 10 repetitions of bird dogs to prepare the core, and 10 dynamic leg swings to prepare the lower body. 

\subsection{Sensor Synchronization}\label{sec:synchronization}
To synchronize the sensors and video recordings, the participants stood in front of the video camera as they listened to an audio recording containing instructions. Participants were instructed to tap on the sensors in time to two different beep trains ,consisting of three beeps to provide the tempo and three beeps to which they tapped along. This was presented at two different tempi to ensure accurate alignment between the video and sensor signals. 

\subsection{Post-climb Questionnaire}\label{sec:anxiety_thermometer}
\begin{figure}[H]
    \centering
    \includegraphics[width=1\linewidth]{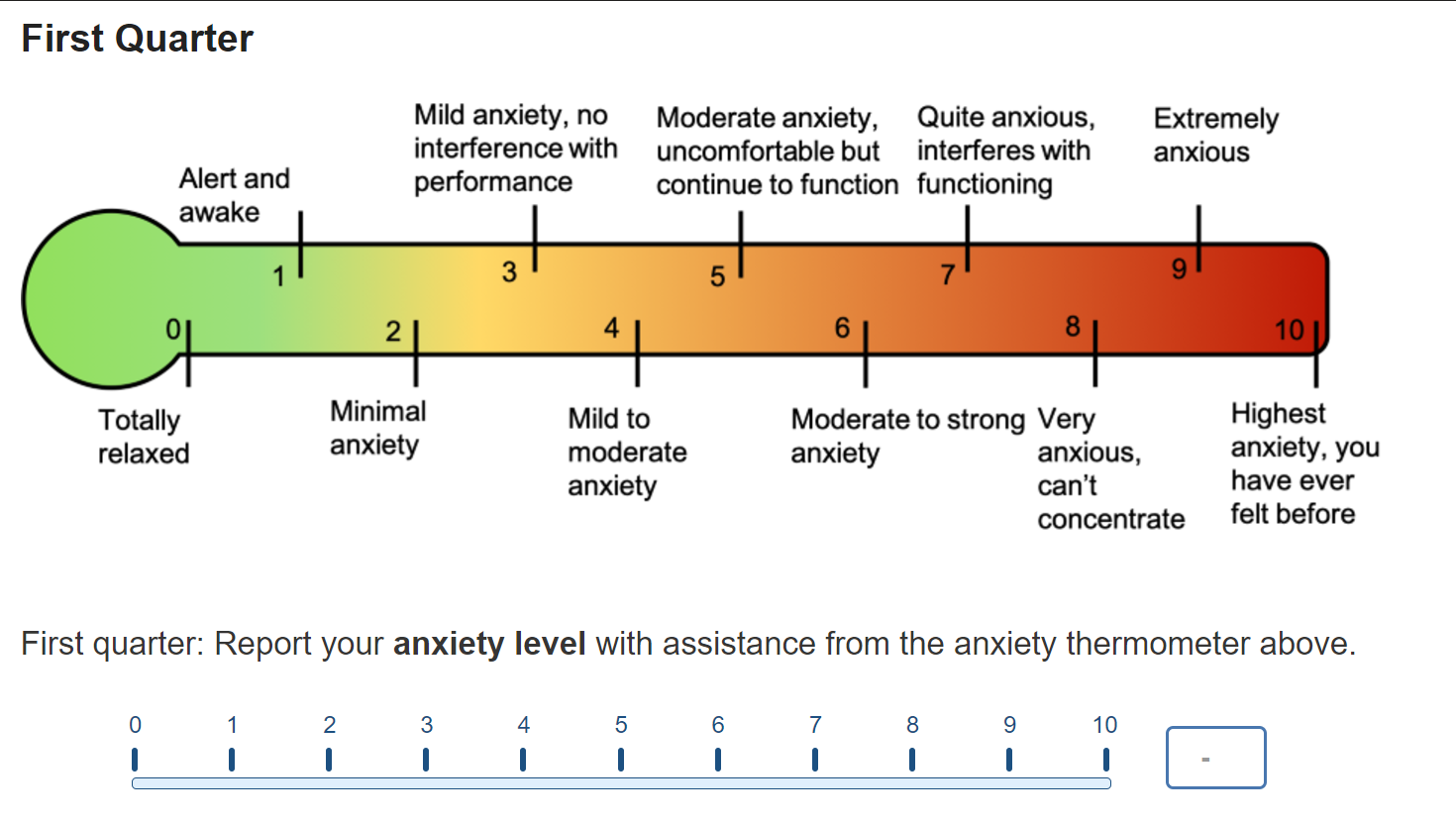}
    \label{fig:anxietyThermometer}
\end{figure}
\begin{figure}[H]
    \centering
    \includegraphics[width=1\linewidth]{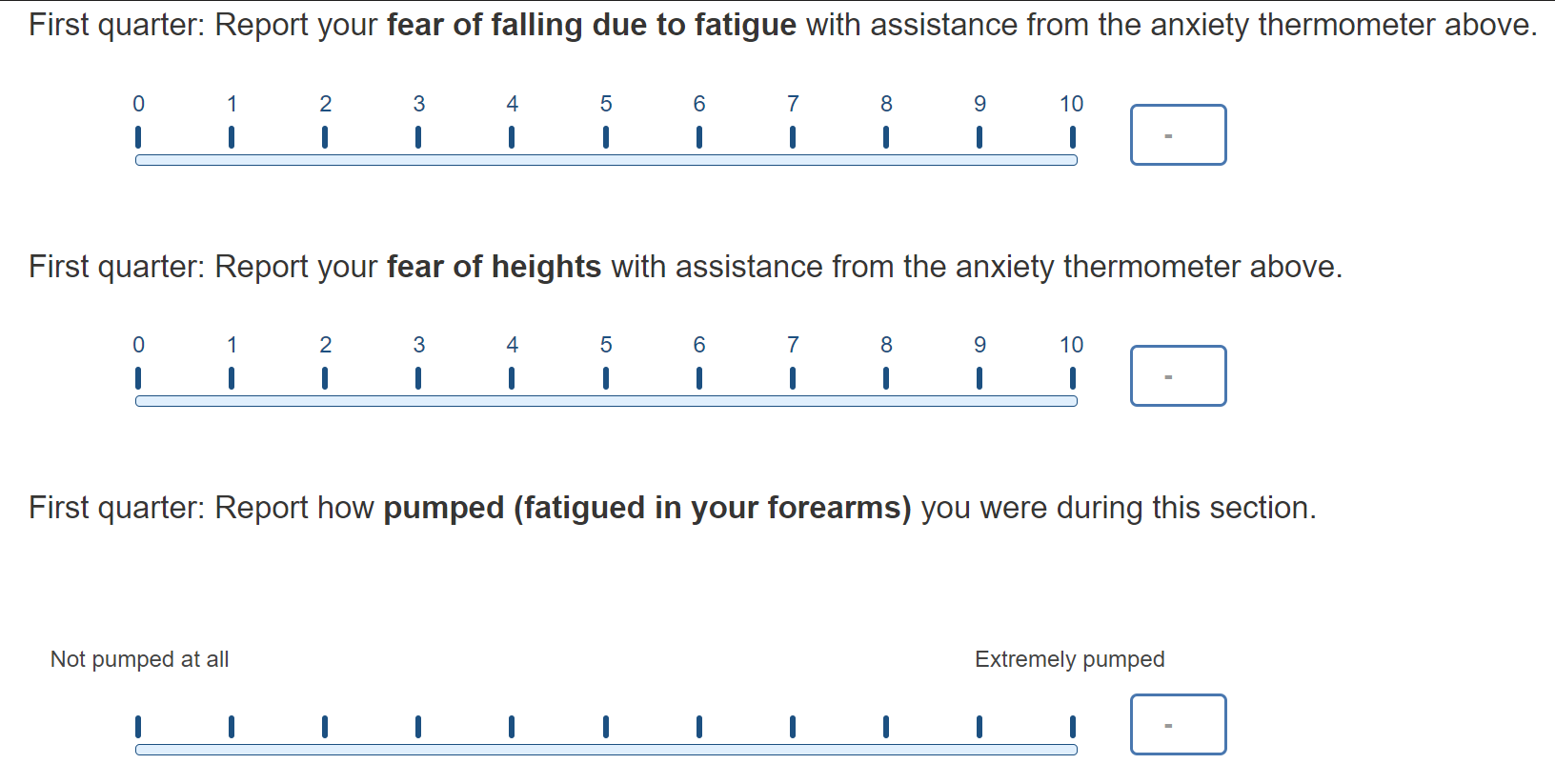}
    \caption{Participants responded to the post-climb questionnaire as they watched the video recording of the climb that they just completed to report their i) anxiety level, ii) fear of falling attributable to fatigue, iii) fear of heights, and iv) fatigue in their forearms}
    \label{fig:enter-label}
\end{figure}
\subsection{Post-experiment Questionnaire}\label{sec:STAI}
\begin{figure}[H]
    \centering
    \includegraphics[width=1\linewidth]{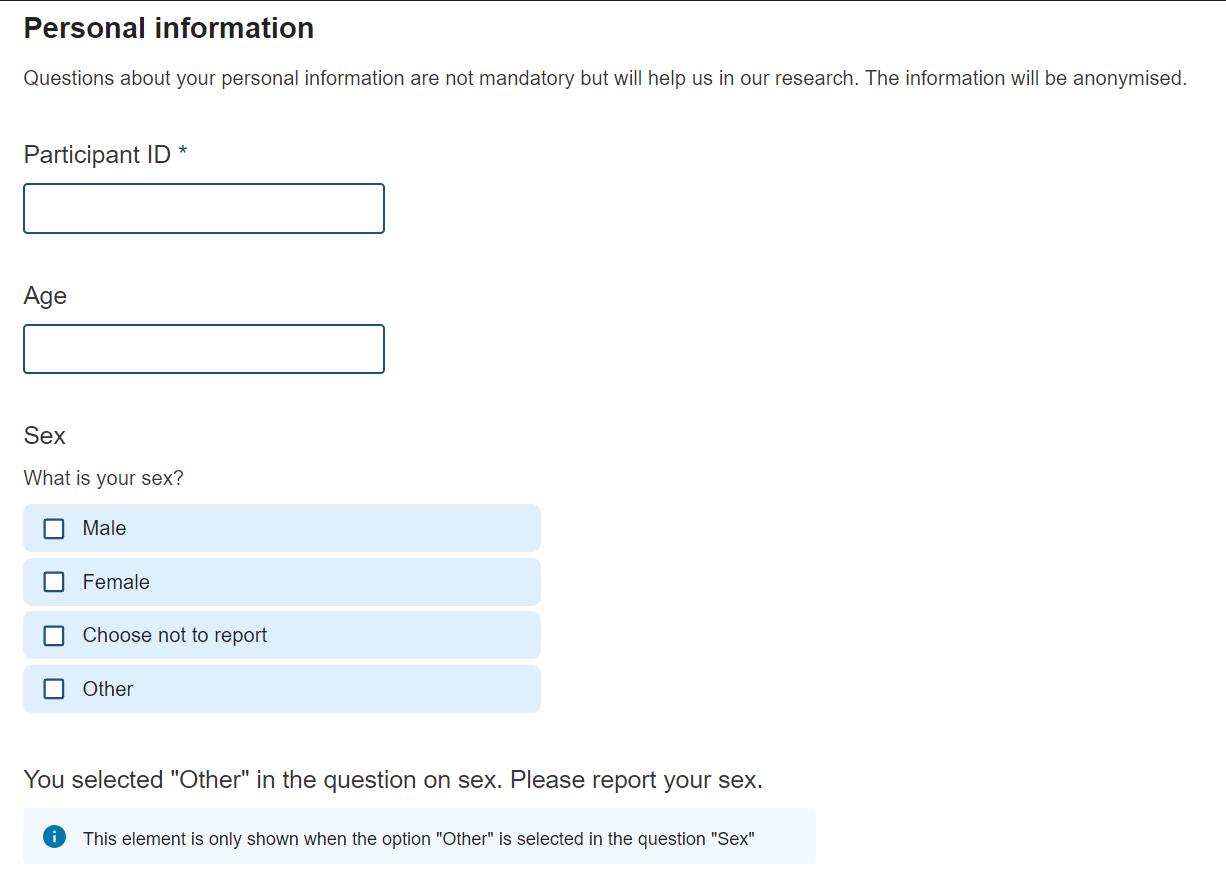}
\end{figure}
\begin{figure}[H]
    \centering
    \includegraphics[width=1\linewidth]{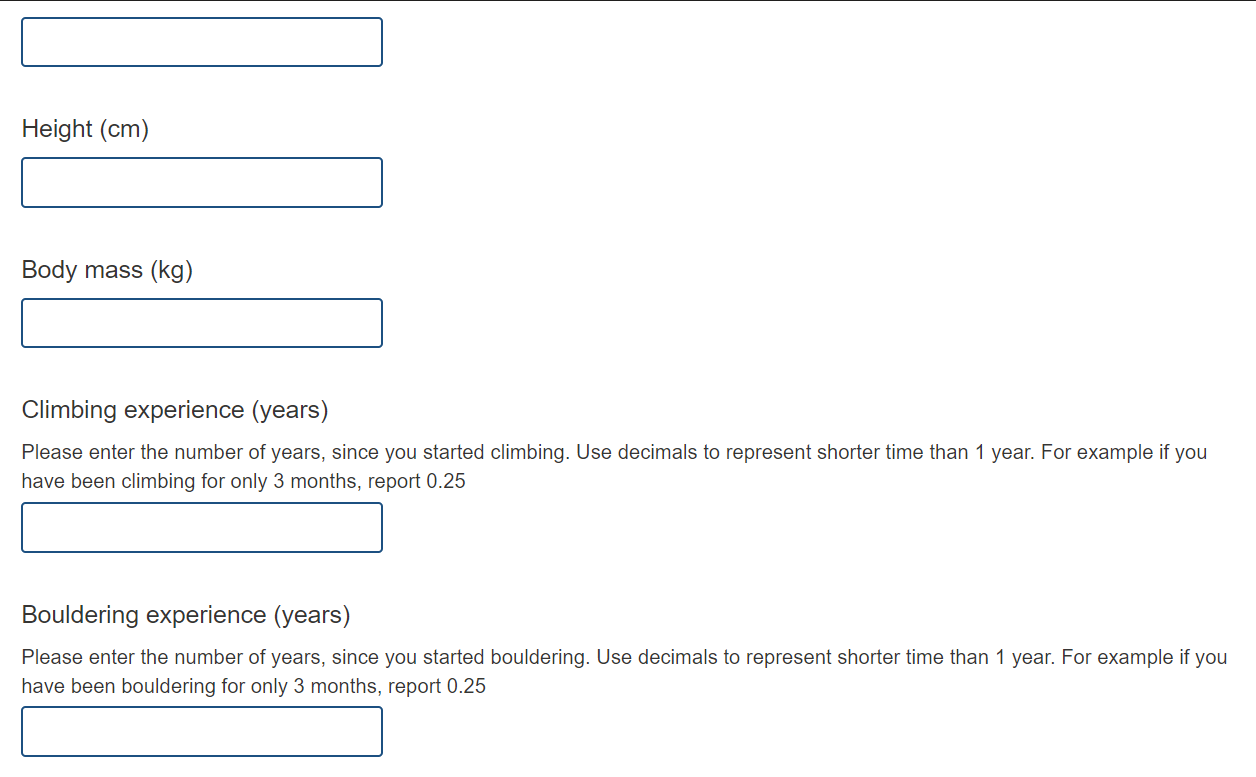}
\end{figure}
\begin{figure}[H]
    \centering
    \includegraphics[width=1\linewidth]{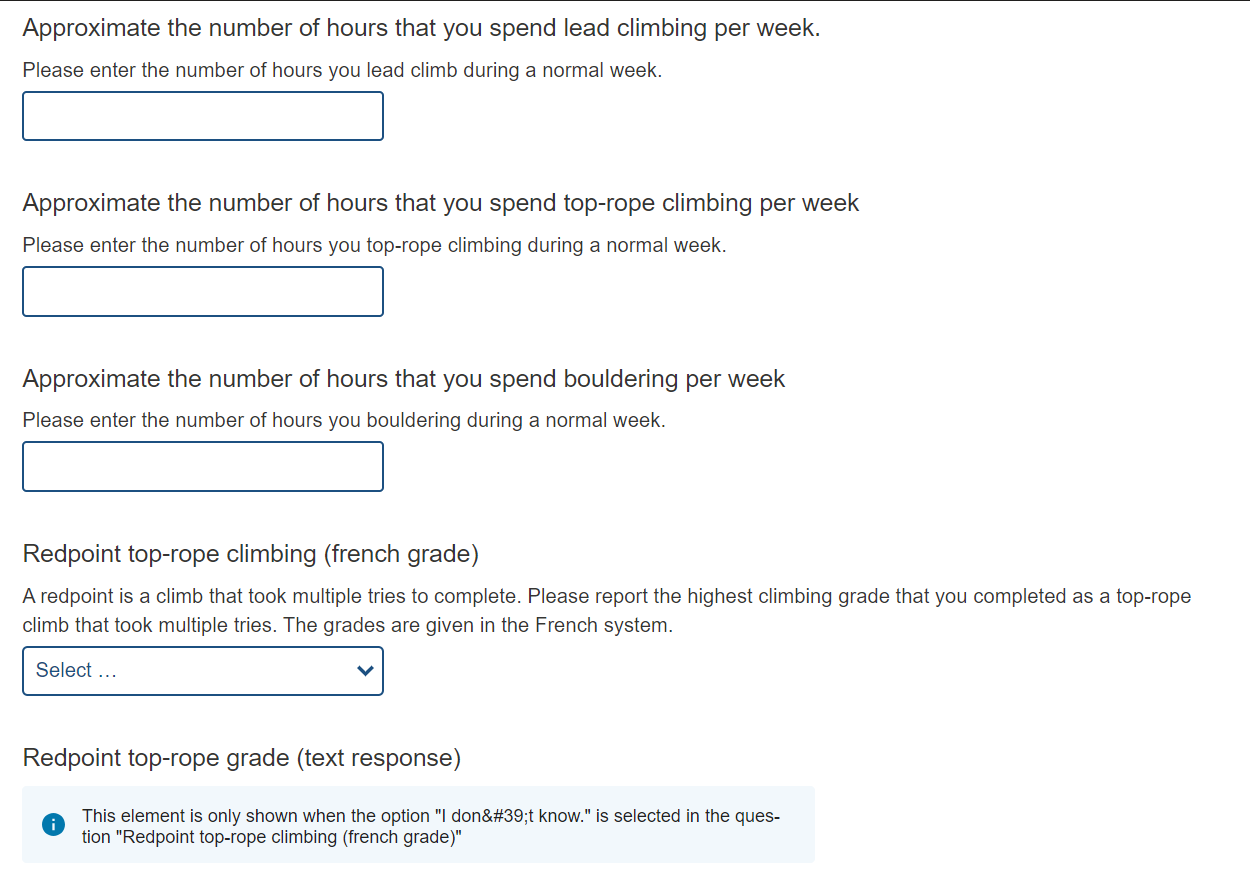}
\end{figure}
\begin{figure}[H]
    \centering
    \includegraphics[width=1\linewidth]{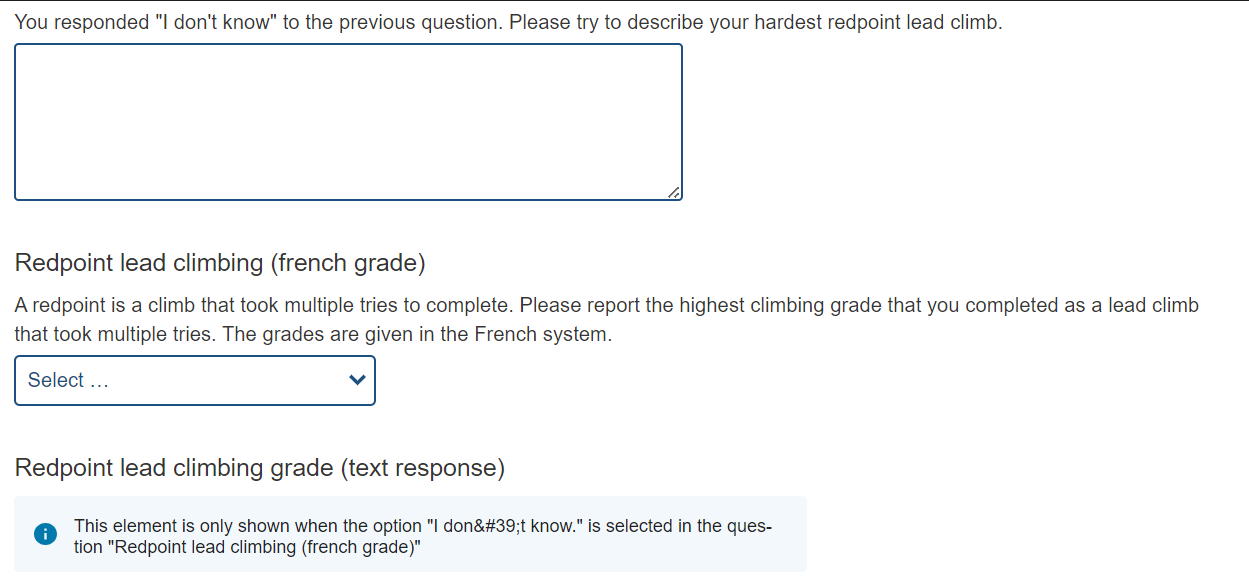}
\end{figure}
\begin{figure}[H]
    \centering
    \includegraphics[width=1\linewidth]{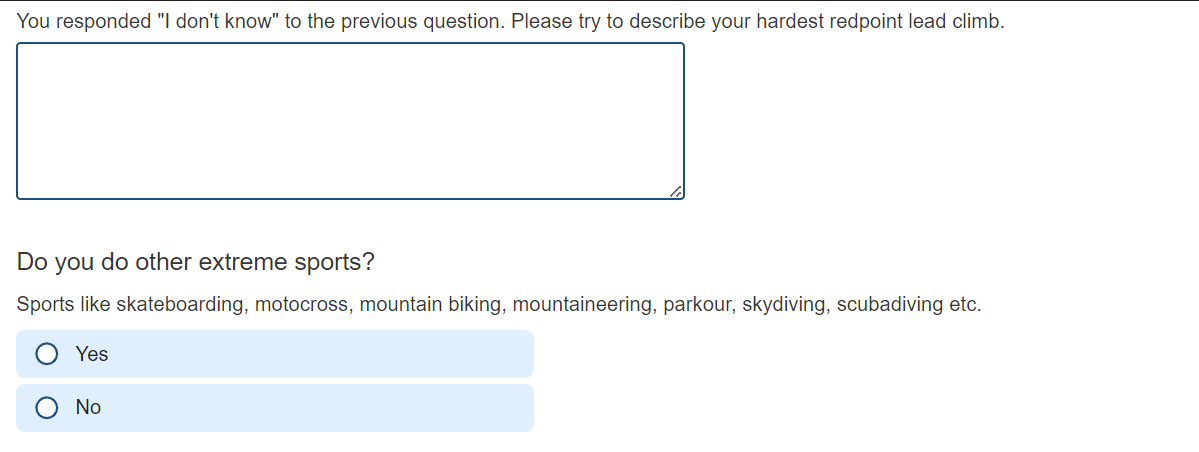}
\end{figure}
\begin{figure}[H]
    \centering
    \includegraphics[width=1\linewidth]{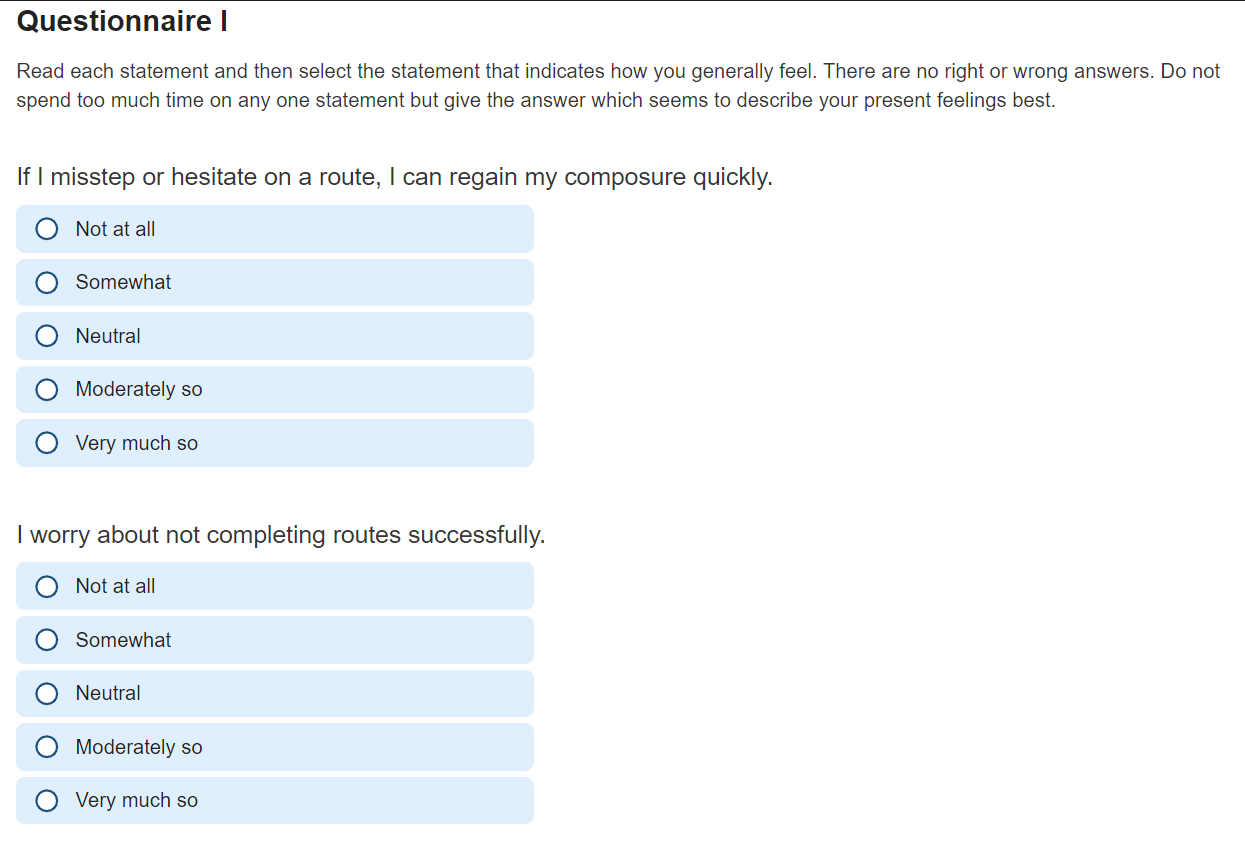}
\end{figure}
\begin{figure}[H]
    \centering
    \includegraphics[width=1\linewidth]{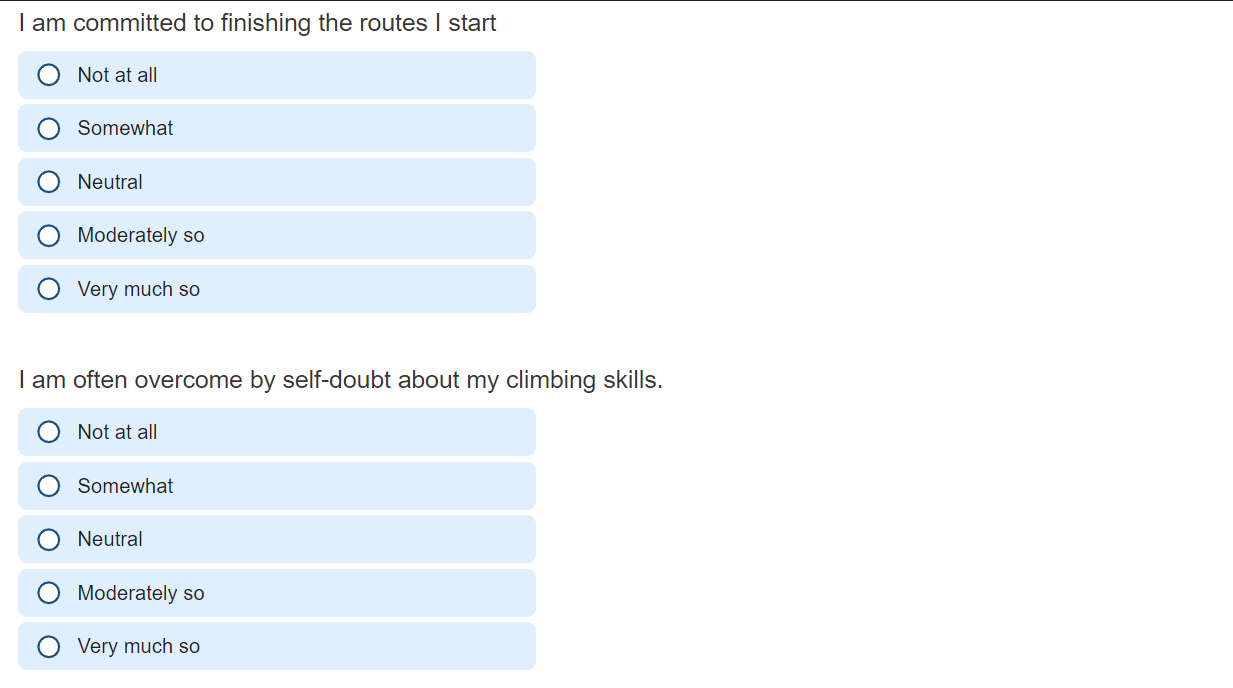}
\end{figure}
\begin{figure}[H]
    \centering
    \includegraphics[width=1\linewidth]{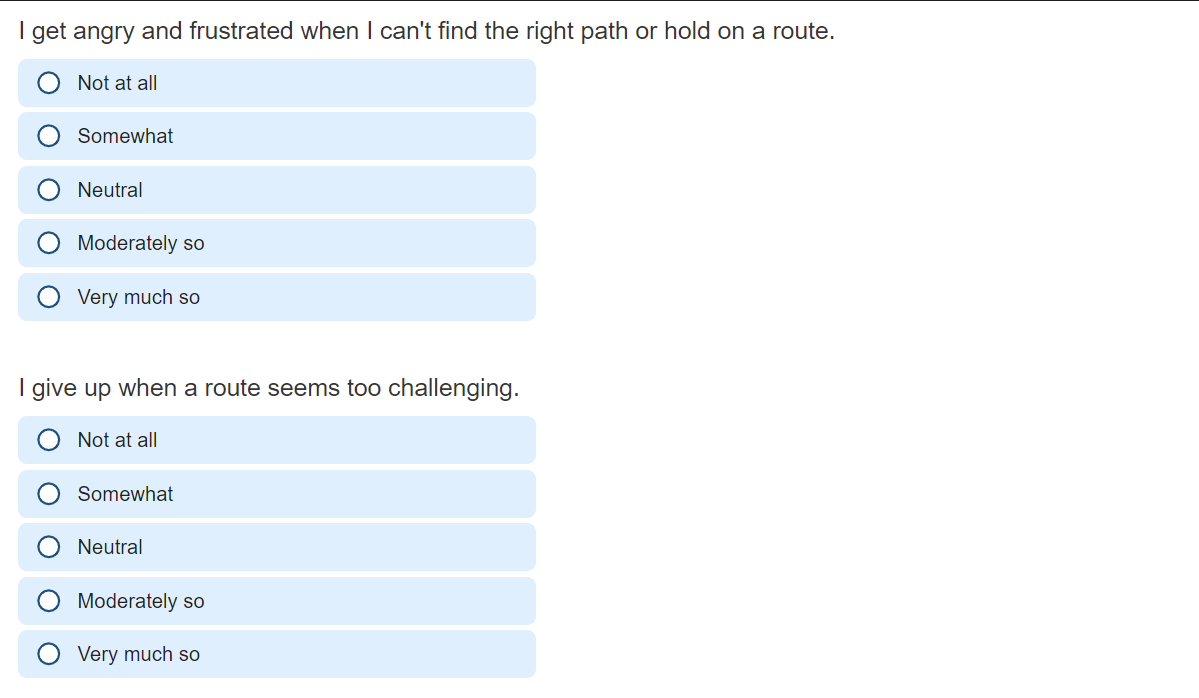}
\end{figure}
\begin{figure}[H]
    \centering
    \includegraphics[width=1\linewidth]{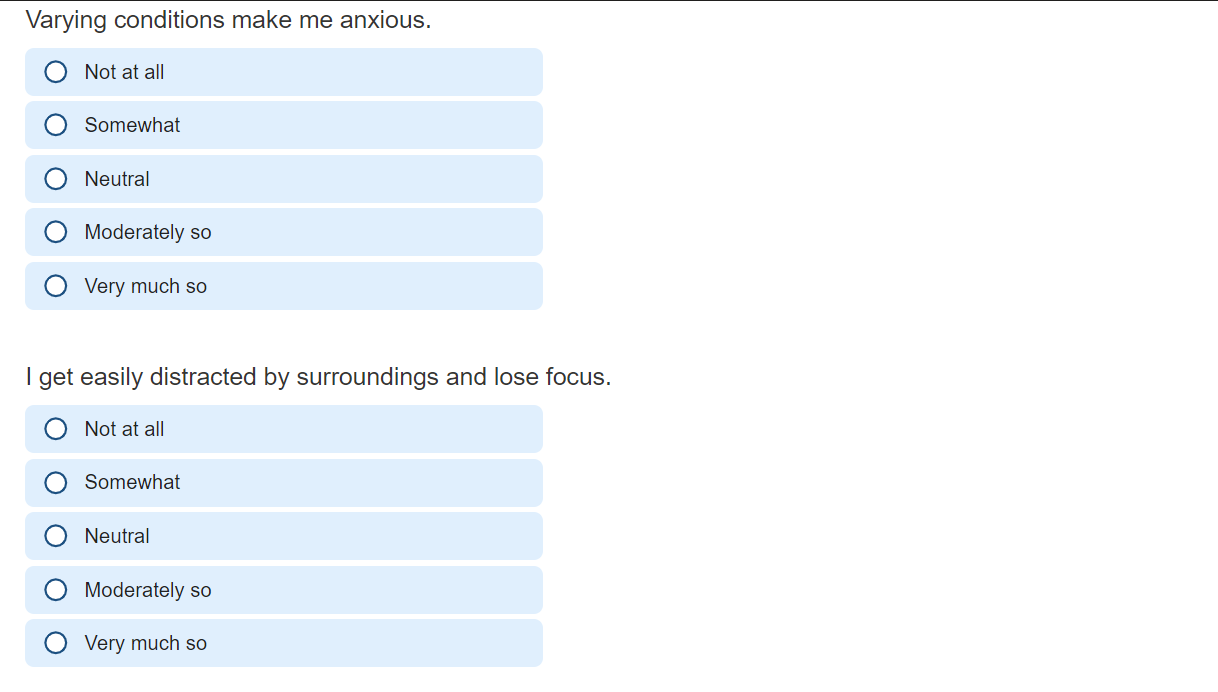}
\end{figure}
\begin{figure}[H]
    \centering
    \includegraphics[width=1\linewidth]{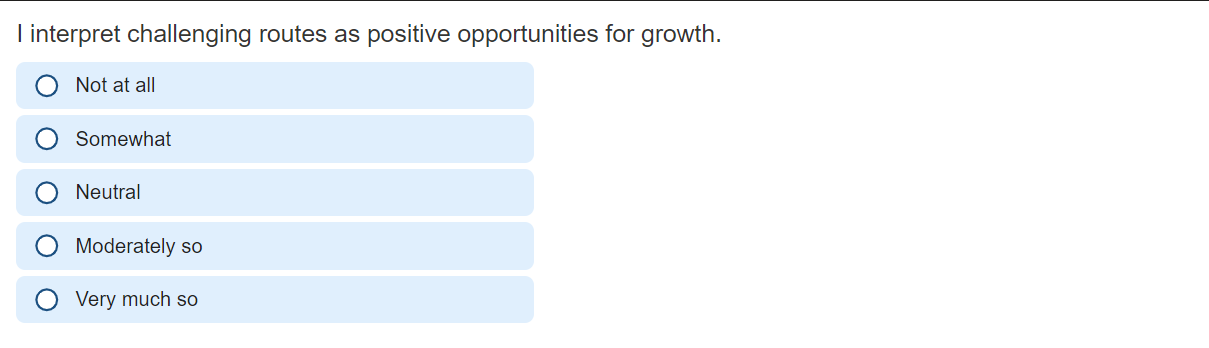}
\end{figure}
\begin{figure}[H]
    \centering
    \includegraphics[width=1\linewidth]{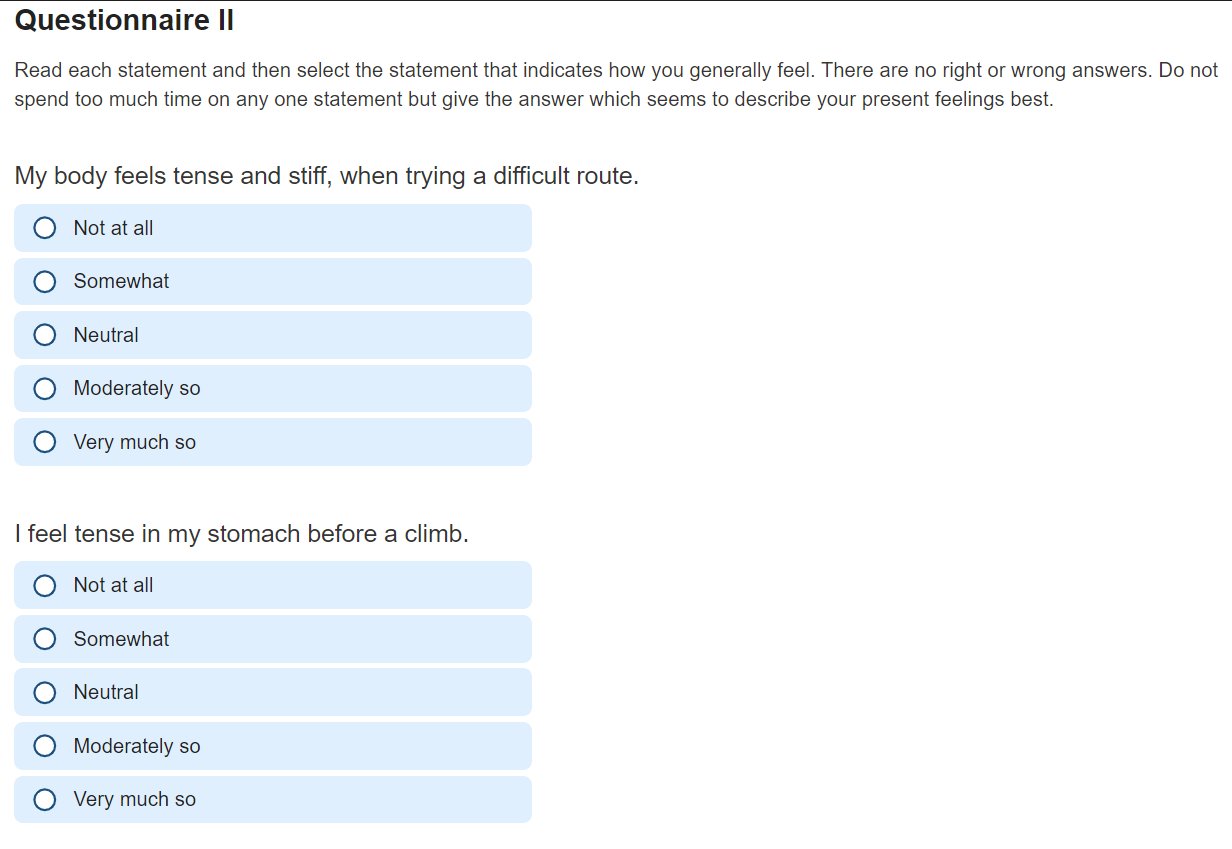}
\end{figure}
\begin{figure}[H]
    \centering
    \includegraphics[width=1\linewidth]{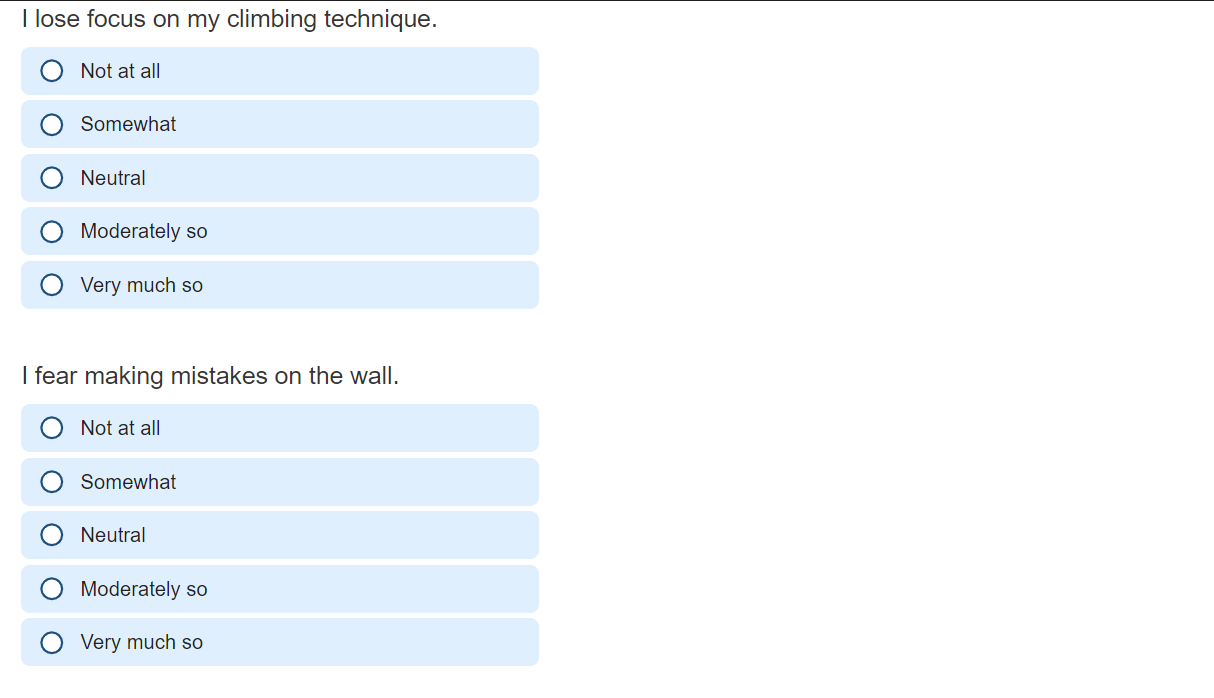}
\end{figure}
\begin{figure}[H]
    \centering
    \includegraphics[width=1\linewidth]{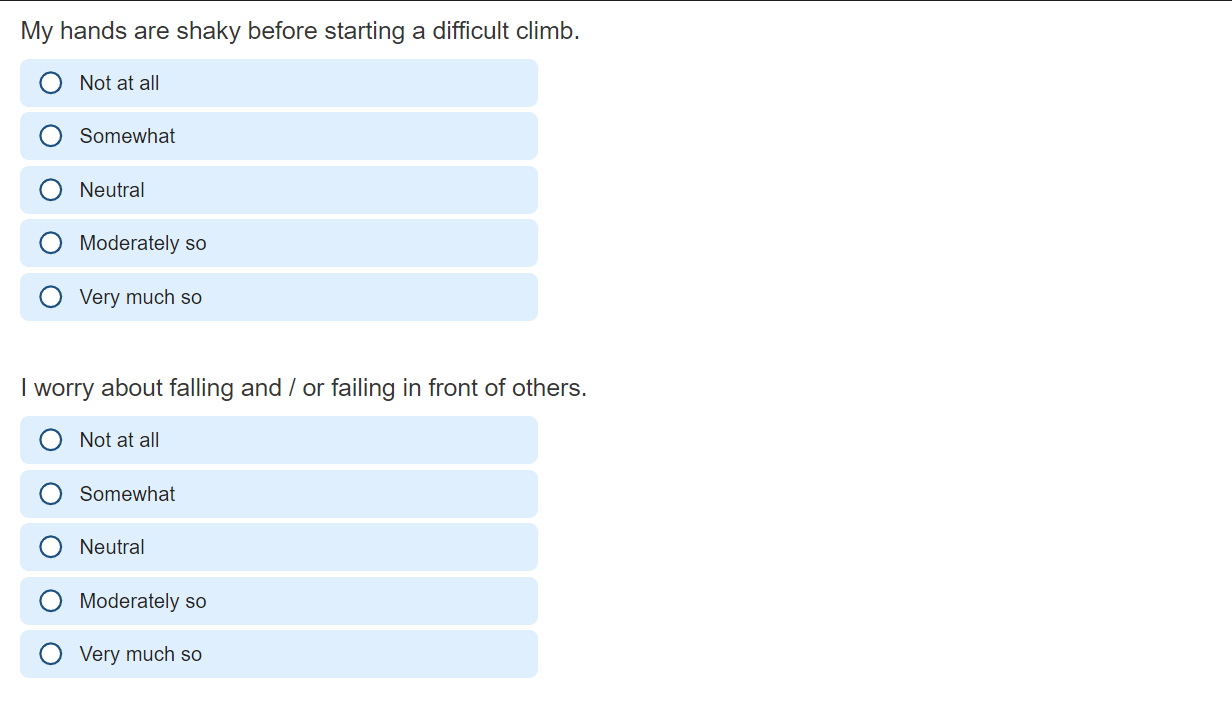}
\end{figure}
\begin{figure}[H]
    \centering
    \includegraphics[width=1\linewidth]{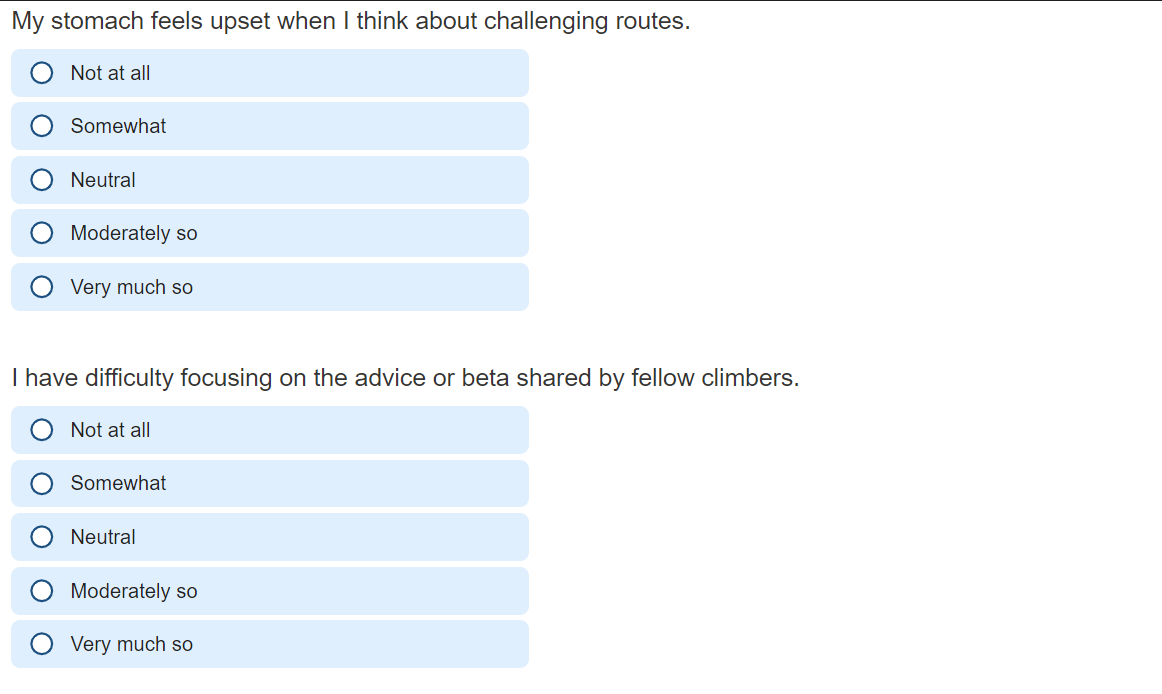}
\end{figure}
\begin{figure}[H]
    \centering
    \includegraphics[width=1\linewidth]{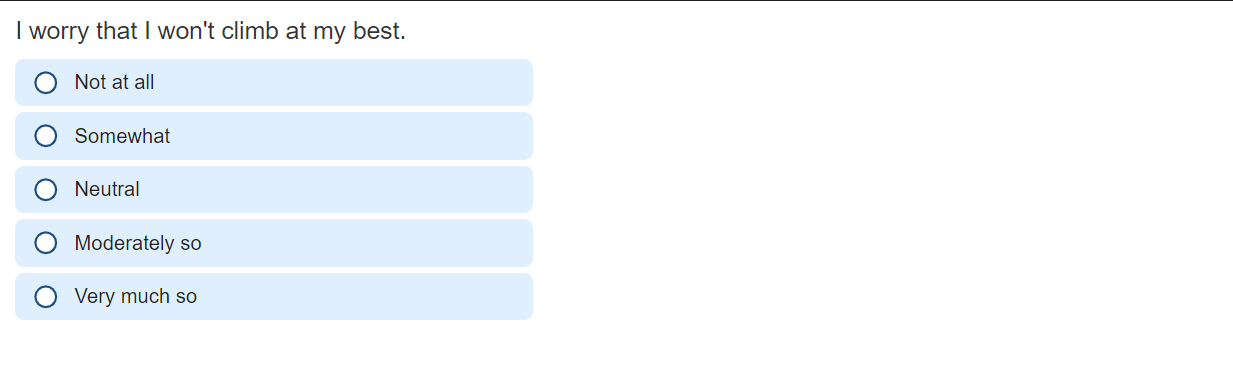}
\end{figure}
\subsection{Validation of linear mixed model assumptions}
\begin{figure}[H]
    \centering
    \includegraphics[width=1\linewidth]{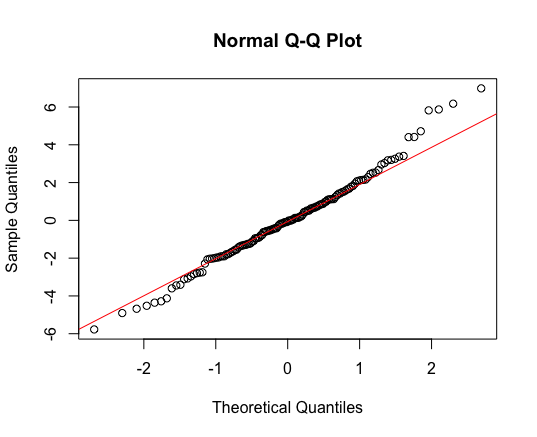}
    \caption{The Q-Q diagram shows how well the residuals of the linear mixed model correspond to a normal distribution. Points that are close to the reference line indicate that the residuals follow a normal distribution. We see that the residuals are mostly aligned with the reference line, which proves that the normality assumption is reasonably fulfilled.}
    \label{fig:qq-diagram}
\end{figure}

\begin{figure}[H]
    \centering
    \includegraphics[width=1\linewidth]{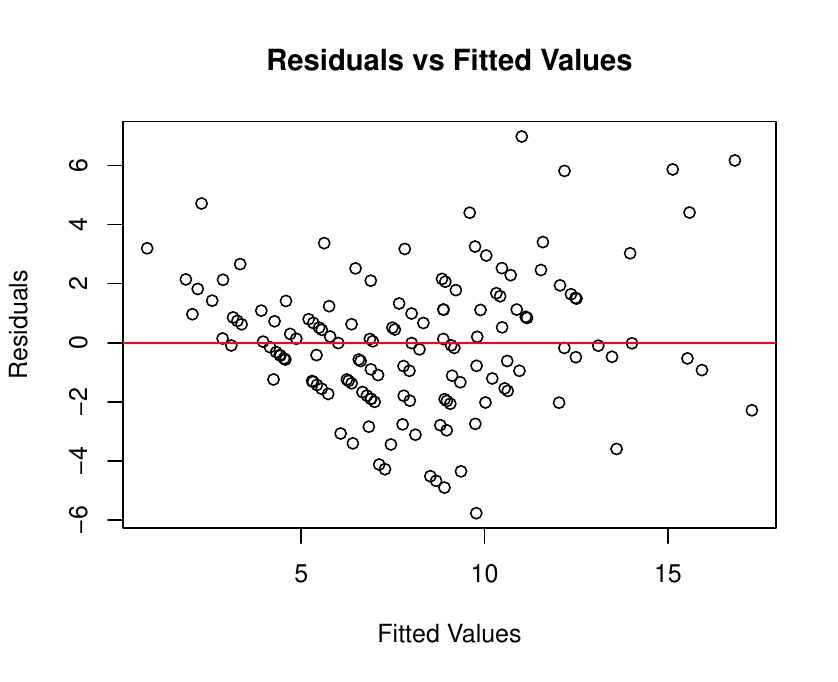}
    \caption{Residuals plotted against fitted values of the regression model. The random dispersion of residuals around zero indicates homoscedasticity and proper model specification.    \label{fig:resid-fitted}}
\end{figure}

\textbf{color}{In addition to the statistical analysis, we provide diagnostics for the linear mixed model to validate its assumptions. First, we evaluated whether the residuals were normally distributed or not. Figure \ref{fig:qq-diagram} shows the Q-Q diagram, which visually evaluates the normal distribution. The residuals appear to follow the normal distribution within the first standard deviation. Therefore, the residuals represented a normal distribution.}
In the next step, we evaluate the assumption of heteroscedasticity. Figure \ref{fig:resid-fitted} shows the residuals plotted against the fitted values of the regression model. In a well-fitted model, the residuals should scatter randomly around zero, indicating homoscedasticity and a correct model specification. The observed random dispersion of the residuals indicates that these assumptions were fulfilled. Deviations from this pattern, such as systematic curves, clusters, or funnel shapes, may indicate problems such as heteroscedasticity, omitted variables, or non-linearity.

\textbf{color}{We used the Goldfeld-Quandt test to test the assumption of homoscedasticity in the residuals of the model. In this test, the residuals were divided into two groups and tested for a significant difference in the variance between them. The test statistic (GQ = 1.155) and the associated p-value (p = 0.296) showed no significant indication of heteroskedasticity. This indicates that the variance of the residuals in the two segments remained constant. }
\subsection{Trainig and validation loss}\label{sec:training_validation_loss}
\begin{figure}
    \centering
    \includegraphics[width=0.5\linewidth]{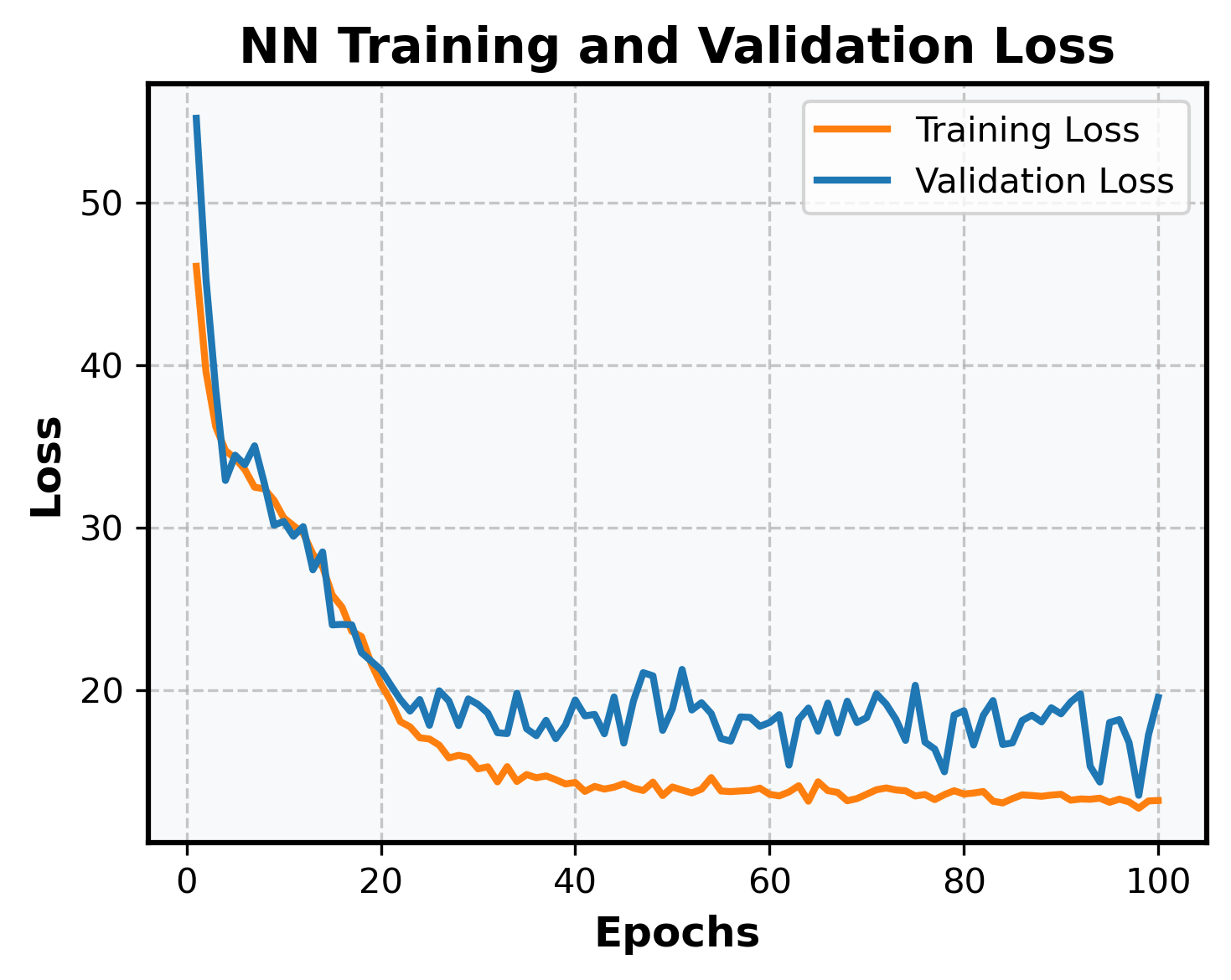}
    \caption{\gls{mlp} training and validation loss.}
    \label{fig:enter-label}
\end{figure}
\begin{figure}
    \centering
    \includegraphics[width=0.5\linewidth]{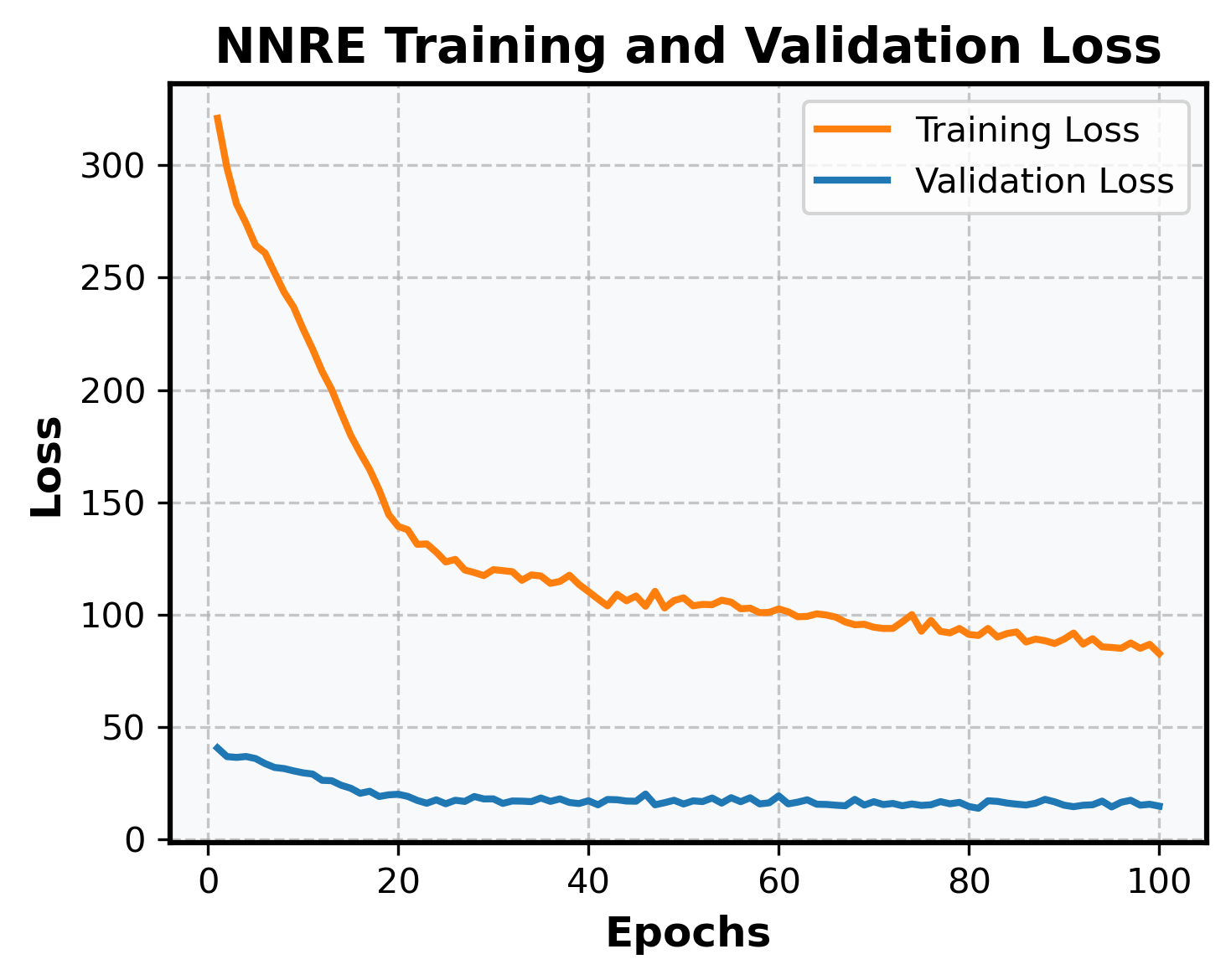}
    \caption{Random effect \gls{mlp} training and validation loss. }
    \label{fig:enter-label}
\end{figure}
\begin{figure}
    \centering
    \includegraphics[width=0.5\linewidth]{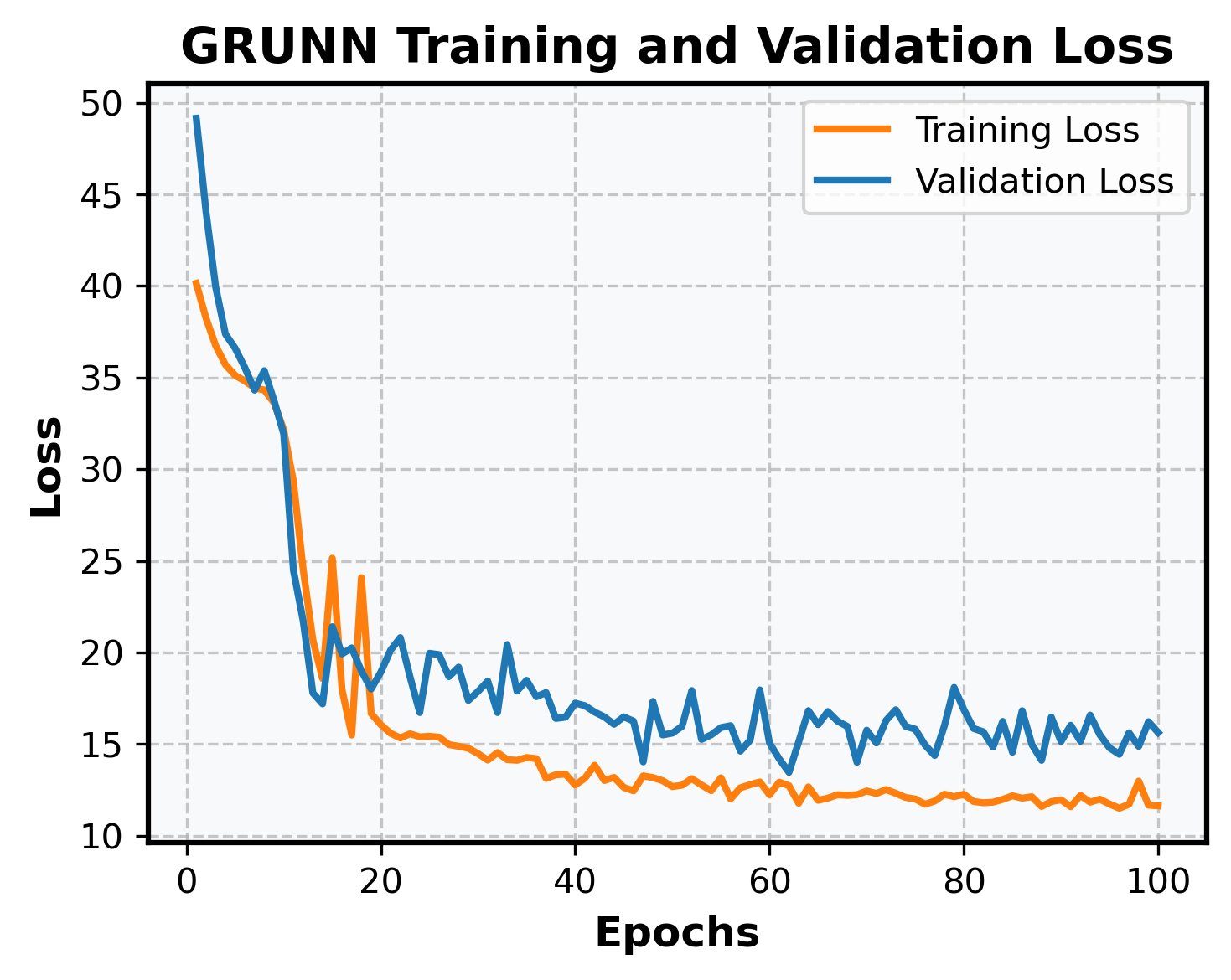}
    \caption{\gls{gru} training and validation loss.}
    \label{fig:enter-label}
\end{figure}
\begin{figure}
    \centering
    \includegraphics[width=0.5\linewidth]{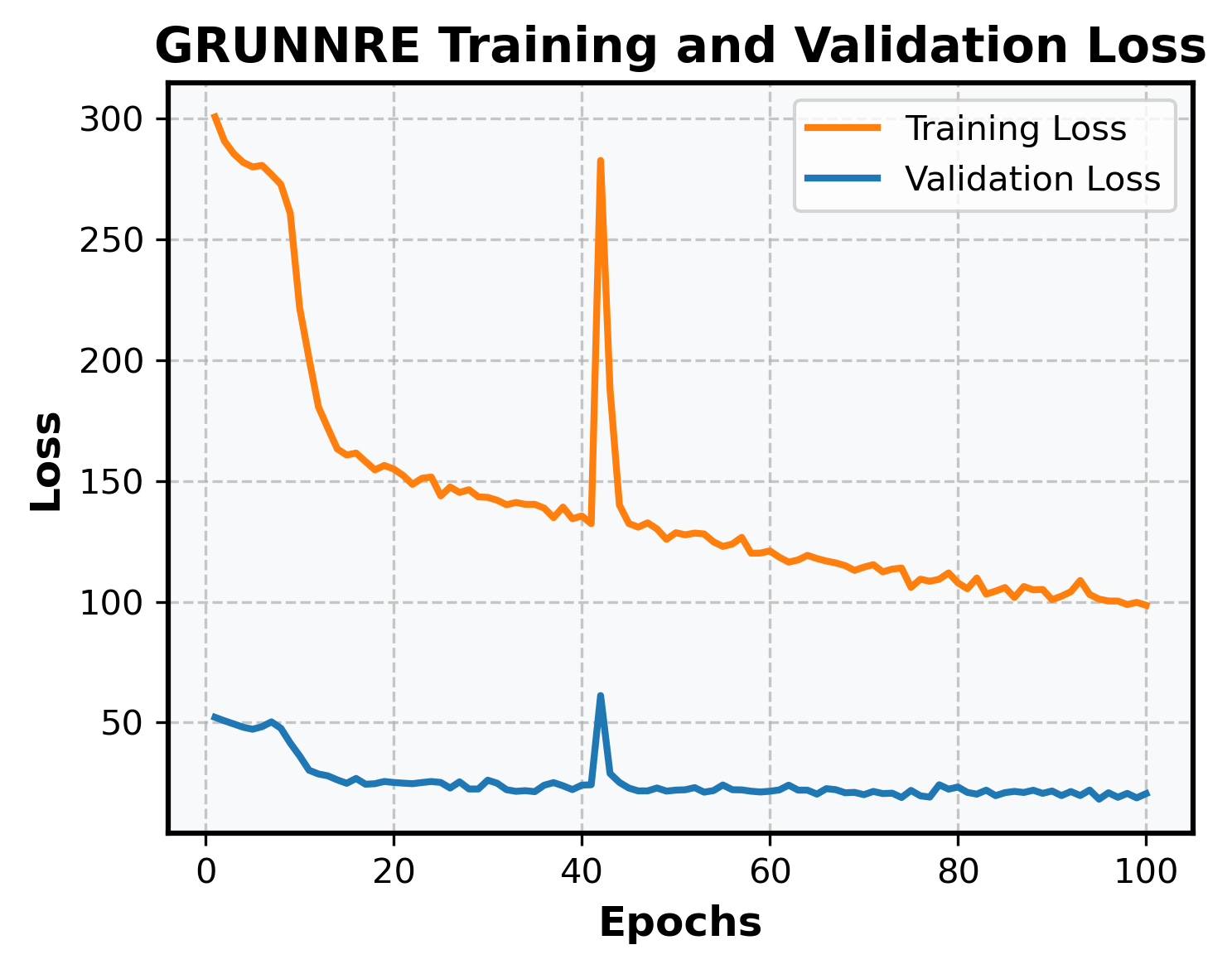}
    \caption{Random effect \gls{gru} training and validation loss. }
    \label{fig:enter-label}
\end{figure}
\bibliographystyle{unsrt}
\bibliography{references}
\end{document}